%% file: main.tex
\documentclass{article}

\usepackage[preprint]{neurips_2021}

\usepackage[utf8]{inputenc} 
\usepackage[T1]{fontenc}    
\usepackage{url}            
\usepackage{booktabs}       
\usepackage{amsfonts}       
\usepackage{amsmath}
\usepackage{nicefrac}       
\usepackage{microtype}      
\usepackage{graphicx}
\usepackage{placeins}
\usepackage{caption}
\usepackage{subcaption}
\usepackage[dvipsnames]{xcolor}
\usepackage{acronym}
\usepackage{subcaption}
\usepackage{hyperref}       
\usepackage{epstopdf}
\input{acronyms.tex}

\hypersetup{colorlinks=true, breaklinks=true, pagebackref=true,
  urlcolor=blue, linkcolor=blue,anchorcolor=blue,citecolor=blue,
  pdfpagemode = UseNone, 
  pdfauthor = {},
  pdftitle = {},   
  pdfsubject = {},
  pdfkeywords = {}
}

\makeatletter
\newcommand*{\centerfloat}{%
  \parindent \z@
  \leftskip \z@ \@plus 1fil \@minus \textwidth
  \rightskip\leftskip
  \parfillskip \z@skip}
\makeatother

\title{Online Training of Spiking Recurrent Neural Networks with Phase-Change
  Memory Synapses}

\author{%
  Yi\u{g}it Demira\u{g}\\
  The Institute of Neuroinformatics\\
  University of Zurich, ETH Zurich\\
  Zürich, Switzerland \\
  \texttt{yigit@ini.ethz.ch} \\
  \And
  Charlotte Frenkel \\
  The Institute of Neuroinformatics\\
  University of Zurich, ETH Zurich\\
  Zürich, Switzerland \\
  \And
   Melika Payvand \\
   The Institute of Neuroinformatics\\
   University of Zurich, ETH Zurich\\
  Zürich, Switzerland \\
  \And
  Giacomo Indiveri \\
  The Institute of Neuroinformatics\\
  University of Zurich, ETH Zurich\\
  Zürich, Switzerland \\
}

\begin{document}
\maketitle
\begin{abstract}
  Spiking \acp{RNN} are a promising tool for solving a wide variety of complex cognitive and motor tasks, due to their rich temporal dynamics and sparse processing.
  However training spiking \acp{RNN} on dedicated neuromorphic hardware is still an open challenge.
  This is due mainly to the lack of local, hardware-friendly learning mechanisms that can solve the temporal credit assignment problem and ensure stable network dynamics, even when the weight resolution is limited.
  These challenges are further accentuated, if one resorts to using memristive devices for in-memory computing to resolve the von-Neumann bottleneck problem, at the expense of a substantial increase in variability in both the computation and the working memory of the spiking \acp{RNN}.
  To address these challenges and enable online learning in memristive neuromorphic \acp{RNN}, we present a simulation framework of differential-architecture crossbar arrays based on an accurate and comprehensive \ac{PCM} device model.
  We train a spiking \ac{RNN} whose weights are emulated in the presented simulation framework, using a recently proposed e-prop learning rule.
  Although e-prop locally approximates the ideal synaptic updates, it is difficult to implement the updates on the memristive substrate due to substantial \ac{PCM} non-idealities.
  We compare several widely adapted weight update schemes that primarily aim to cope with these device non-idealities and demonstrate that accumulating gradients can enable online and efficient training of spiking \acp{RNN} on memristive substrates.
\end{abstract}

\section{Introduction}
\acp{RNN} are exceptionally expressive~\cite{Khrulkov_etal17} class of neural networks that have been successfully adapted in many domains such as audio processing, optical flow, language modeling and \ac{RL}~\cite{Oord_etal16, Teed_Deng20, Brown_etal20, Berner_etal19, Ha_Schmidhuber18, Vinyals_etal19}.
The power of \ac{RNN}s lie in their architecture that allows processing of long and complex sequential data.
Due to its recurrent architecture, each neuron contribute to the network processing at various points of the computation.
The resulting efficiency of \acp{RNN} is also evident in the mammalian brain with massive lateral recurrent connections in neocortex~\cite{Douglas_Martin98, Douglas_etal95a}.
However, the training of \ac{RNN} topologies is notably difficult under constrained memory and computational resources~\cite{Zenke_Neftci21}.

Current hardware implementations of neural networks still cannot compete with the energy efficiency of biological systems.
One of the main reasons for this difference is due to the data movement between the separated processing and memory units of the von-Neumann architectures used to implement the artificial neural networks.
Recently, new types of compact nanoscale devices have garnered significant attention for implementing artificial synapses~\cite{Ambrogio_etal18,Li_etal18a,Prezioso_etal15,dalgaty_etal21,cai_etal20,Sebastian_etal19} that can implement in-situ learning and break the von Neumann bottleneck~\cite{Backus78,Indiveri_Liu15}.
These \textit{memristive} devices are particularly promising for use in \ac{SNN} architectures, especially for low-power, sparse and massively parallel spike-based neuromorphic systems, that emulate the principles of computation observed in biological brains~\cite{Payvand_etal19, Chicca_Indiveri20}.
In these systems, the synapses (memory) and neurons (processing units) are arranged in a crossbar architecture as shown in Fig.~\ref{fig:mushroom}a, where memristive devices are placed at the junctions, to store the synaptic efficacy in their programmable conductance values.
The crossbar architecture implements \textit{in-memory} computation of synaptic propagation between neurons by a single physics-based operation following Ohm's Law and Kirchhoff's Law, intrinsically supporting the sparse and event-driven nature of \acp{SNN}. 
As recently shown in \cite{Peng_etal19} and \cite{Peng_etal20} for 32\,nm technology, compared to a digital \ac{CMOS} implementation (i.e. \ac{SRAM}) such memristive crossbar arrays enable a denser solution, and have respectively lower and similar dynamic energy consumption during inference and training. 
Moreover, the non-volatile nature of memristive devices reduces the static power consumption usually linked with volatile \ac{CMOS} memory storage.
Therefore, in-memory acceleration of spiking \acp{RNN} with non-volatile multi-bit-resolution memristive devices is a promising direction for scalable neuromorphic hardware for temporal signal processing.

\ac{PCM} devices are among the most mature emerging resistive memory technologies.
Their tiny footprint, fast read/write operation and multi-bit storage capacity make \acp{PCM} an ideal candidate for implementing in-memory computation of synaptic propagation~\cite{Burr_etal16a, Burr_etal17}.
Consequently, there has been an increased interest for the employment of \ac{PCM} technology in neuromorphic computing applications~\cite{Ambrogio_etal18,Tuma_etal16, Karunaratne_etal20a, Demirag_etal21a}.

Single \ac{PCM} device can have 3-4 bits of resolution~\cite{Gallo_etal18}, however, unlike conventional digital \ac{CMOS} memories, they are affected by severe non-idealities as their switching physics is governed by inherently stochastic Joule heating.
Moreover, molecular dynamics of \acp{PCM} give rise to the $1/f$ noise behavior and the structural relaxation, resulting in cycle-to-cycle variation, in addition to their device-to-device variability present due to fabrication effects.

To account for these device non-idealities, one could follow a hardware-algorithm co-design approach with chip-in-the-loop setups that leverage \ac{PCM} crossbar hardware~\cite{Ambrogio_etal18}.
However, training a neural network requires the iterative evaluation of multiple network architectures, modifications to the learning rule and tuning hyperparameters on large datasets, which are extremely time/resource consuming with chip-in-the-loop setups.
On the other hand, a software network simulation framework with a statistical model of memristive devices offers much faster iteration times and a better understanding of device effects on the training, due to increased observability of internal state variables. 
However, it is extremely important to have a very accurate statistical model of the devices simulated, to optimize the training procedure of the network before moving to chip-in-the-loop training.

In this paper, we investigate whether a spiking \ac{RNN} can be trained with a local learning rule despite the adverse impacts of in-memory computing with memristive devices such as write noise, read noise, temporal conductance evolution (i.e., drift) and the limited bit precision.
To do so, we first build on the statistical \ac{PCM} model from Nandakumar et al.~\cite{Nandakumar_etal18} to faithfully model a crossbar array comprising a differential memristor configuration (Section \ref{section:model}).
Then, we define a target spiking \ac{RNN} architecture and describe the properties of an ideal learning rule and select e-prop algorithm~\cite{Bellec_etal20} to train the network (Section \ref{section:architecture}).
In order to map ideal weight updates calculated by e-prop to memristor conductances on the crossbar array, we implement multiple memristor-aware weight update methods that are optimized to cope with device non-idealities (Section \ref{section:updates}). 
Finally, we report a training scheme for spiking \acp{RNN} which exploits in-memory computing with extremely sparse activity and a reduced number of conductance updates for energy efficient training (Section \ref{section:results}).

\newpage
\section{Building Blocks for In-Memory Acceleration of \ac{RNN} Training for Neuromorphic Processors}
In the following, we describe the main components of our simulation framework for training spiking \acp{RNN} with \ac{PCM} synapses~\footnote{The code is available at \url{https://github.com/YigitDemirag/srnn-pcm}}.

\subsection{\ac{PCM} Synapses}
\label{section:model}
\begin{figure}
  \centering
  \includegraphics[width=0.99\textwidth]{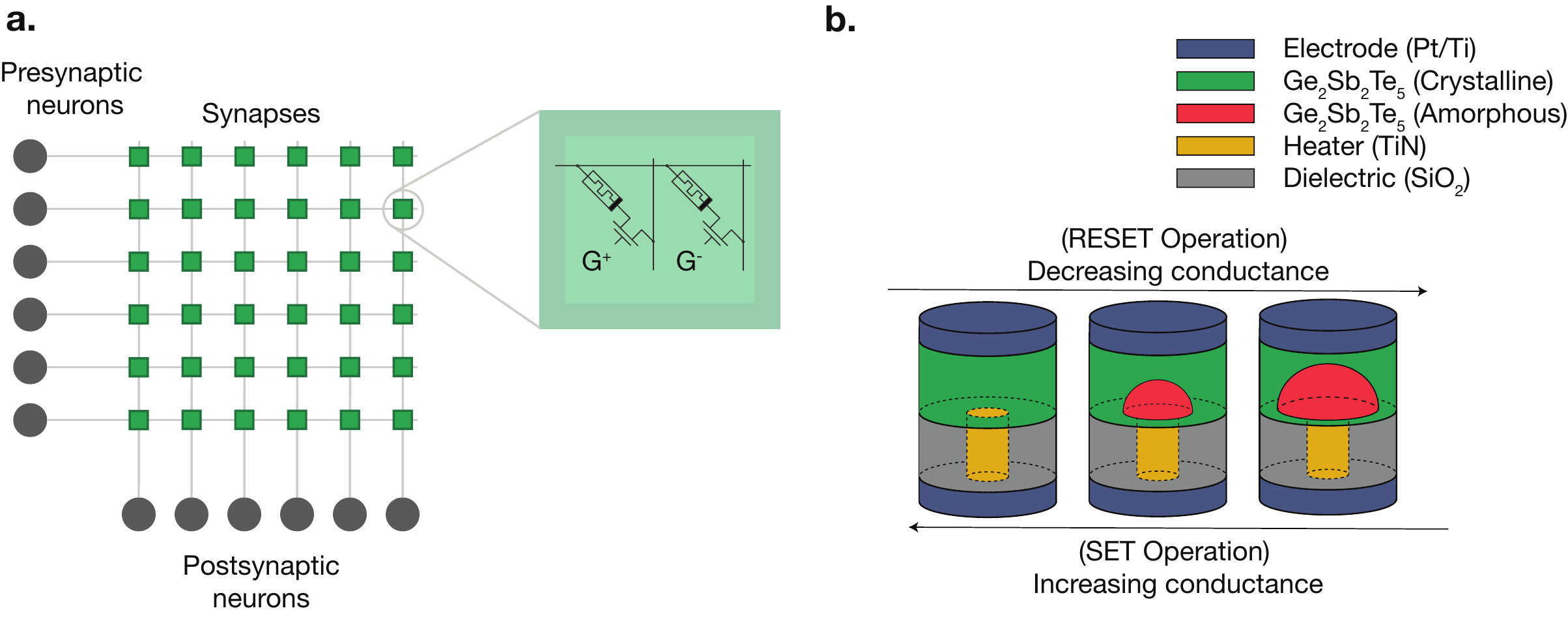}
  \caption{\textbf{a.} \ac{PCM} devices can be arranged in a crossbar
    architecture to emulate both a non-volatile synaptic memory and a parallel and
    asynchronous synaptic propagation using in-memory computation. \textbf{b.}
    Mushroom-type geometry of a single \ac{PCM} device. The conductance
  of the device can be reconfigured by changing the volume ratio of amorphous and
  crystalline regions.}
  \label{fig:mushroom}
\end{figure}

The material design of a nanoscale \ac{PCM} device typically includes the switching material \ac{GST} placed between two metal electrode layers forming a mushroom-type structure (Fig~\ref{fig:mushroom}b).
By applying short electrical pulses, the temperature distribution inside the \ac{PCM} device can be momentarily modified via Joule heating.
The controlled temperature levels can switch the molecular configuration of \ac{GST} between the amorphous (high-resistance) and crystalline (low-resistance) states~\cite{Demirag18}.

To increase the volume of the amorphous state in the device, a short and high-amplitude electrical pulse (RESET pulse) is applied to the device terminals.
Increasing the temperature to around 900 K melts a significant portion of the \ac{GST}, and if the \ac{GST} quenches rapidly, melted region forms an amorphous configuration.
On the contrary, to increase the crystalline volume, typically a longer and smaller-amplitude electrical pulse (SET pulse) is applied to the device terminals.
In this case, temperature rises to around 400-600 K, which initiates the growth of available crystal nuclei inside the \ac{GST} and increases the crystalline volume.
To read the device conductance, an electrical pulse with an even smaller amplitude (READ pulse) is applied to the device, in order to prevent any phase transition.
The amplitude and duration of SET, RESET and READ pulses depend on the \ac{GST} composition and device volume.
Nevertheless for a typical mushroom type geometry with 100 nm \ac{GST} and 20 nm heater radius (see Fig.~\ref{fig:mushroom}b), a SET pulse of 2.5V amplitude and 100 ns duration, a RESET pulse of 3.5 V amplitude and 20 ns duration and READ pulse of 0.2 V 50 ns duration can be used~\cite{Demirag18}.

\FloatBarrier

In practice however, the programming operations of such devices suffer from write noise, read noise, and the electrical conductance drift~\cite{Gallo_Sebastian20}.
The asymmetry of the SET and RESET operations, and the nonlinear conductance response with respect to the number and frequency of the pulses applied further complicates the precise programming of the device conductance to target values.
It is therefore critical to fully capture these non-ideal phenomena and device dynamics in the network models, because only then important metrics such as the robustness of the weight update rule, hyperparameter choices or the training duration can be realistically evaluated, reflecting the real-world deployment of the neuromorphic system.
Many comprehensive models have been proposed to describe electrical \cite{Gallo_etal15}, thermal \cite{Gallo_etal16}, structural \cite{Ielmini_etal07, Karpov_etal07} and phase-change \cite{Redaelli_etal08, Salinga_etal13} properties of \acp{PCM}.
These device models either require solving on-the-fly differential equations whose numerical convergence is not guaranteed, or do not incorporate inter- and intra- device stochasticity, or are designed for pulse shapes and current-voltage sweeps that do not reflect the operational conditions on the circuit~\cite{Demirag18}.

\begin{figure}
  \centerfloat
  \includegraphics[width=1.15\textwidth]{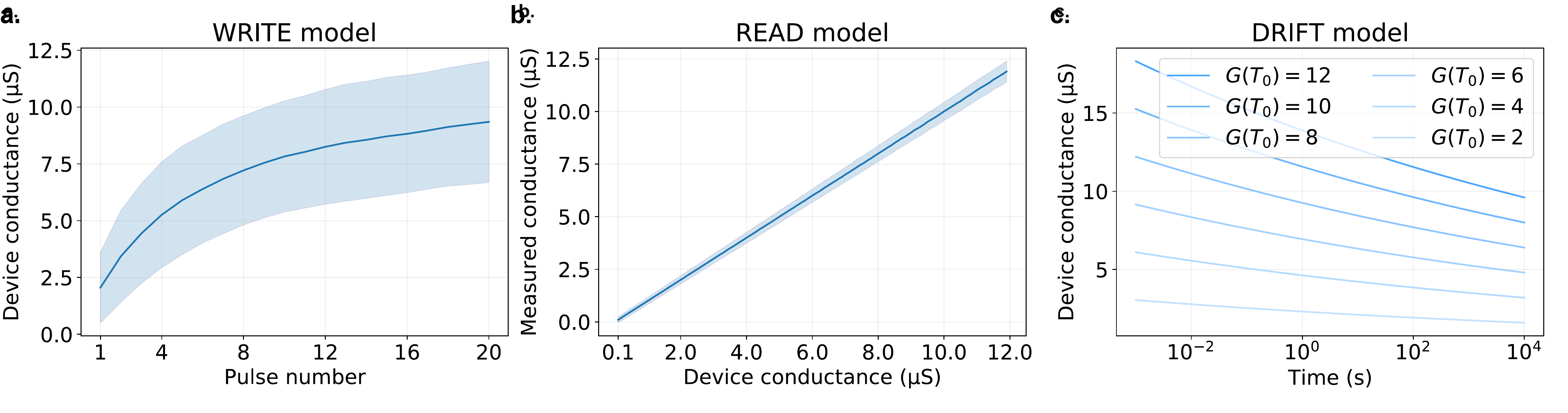}
  \caption{The chosen \ac{PCM} model from~\cite{Nandakumar_etal18} captures the major device non-idealities.
  \textbf{a.} The WRITE model enables calculation of the
    conductance increase with each consecutive SET pulse applied to the device. The
    band illustrates one standard deviation. \textbf{b.} The READ model enables
    calculation of $1/f$ noise, increasing as a function of conductance.
    \textbf{c.} The DRIFT model calculates the temporal conductance evolution as a
    function time. $T_0$ indicates the time of measurement after the initial
    programming of the device.}
  \label{fig:model}
\end{figure}

In this study, we selected the statistical device model from Nandakumar et al.~\cite{Nandakumar_etal18}, which comprises all major forms of \ac{PCM}-specific non-idealities based on measurements from 10,000 devices.
The model includes the nonlinear conductance change with respect to applied pulses, conductance-dependent write and read operation stochasticity, and the temporal drift effect (Fig.~\ref{fig:model}).
The model keeps a programming history variable, which represents the nonlinear device response to consecutive SET pulses and is updated after the application of each pulse.
Following the application of a new SET pulse, the model samples the conductance change $\Delta G$ from a Gaussian distribution whose mean and standard deviation is based on the programming history and the previous conductance.
The drift behavior is then included following the empirical exponential drift model \cite{Karpov_etal07} $G(t)=G\left(T_{0}\right)\left(t/T_{0}\right)^{-v}$, where $G(T_0)$ is the estimated conductance after a WRITE pulse at time $T_0$ and $G(t)$ is the final device conductance considering the effect of the drift.
Additionally, the model takes into account the $1/f$ READ noise~\cite{Nardone_etal09}, which increases monotonically with the device conductance.
Overall, this statistical \ac{PCM} model captures the stochastic conductance changes due to the application of SET and READ pulses and estimates the temporal conductance drift arising from the structural relaxation.

In order to integrate this model into neural network simulations, we developed a comprehensive \ac{PCM} crossbar array simulation framework in PyTorch~\cite{Paszke_etal19a}.
Our crossbar array simulation framework can keep track of all simulated \ac{PCM} devices simultaneously, enabling realistic SET, RESET and READ operations (for implementation details, see \nameref{section:xbar}).
Section~\ref{section:trainingsnn} will describe how this framework is used to represent synaptic weights of a \ac{RNN}.

\FloatBarrier

\subsection{Credit Assignment Solutions for Recurrent Network Architectures}
\label{section:architecture}
The credit assignment problem refers to the problem of determining the amount of change required in each synaptic weight to achieve the network's  desired behavior~\cite{Richards_etal19}.
The nature of this problem is intertwined with network architecture which describes the arrangement of synapses, neurons and their operational principles.
Hence many proposed solutions to the credit assignment problem in \acp{SNN} landscape are specific to a network architecture components e.g., eligibility traces~\cite{Fremaux_Gerstner16}, dendritic~\cite{Sacramento_etal18} or neuromodulatory~\cite{Pozzi_etal18} signals.

\begin{figure}
  \centering
  \includegraphics[width=1\textwidth]{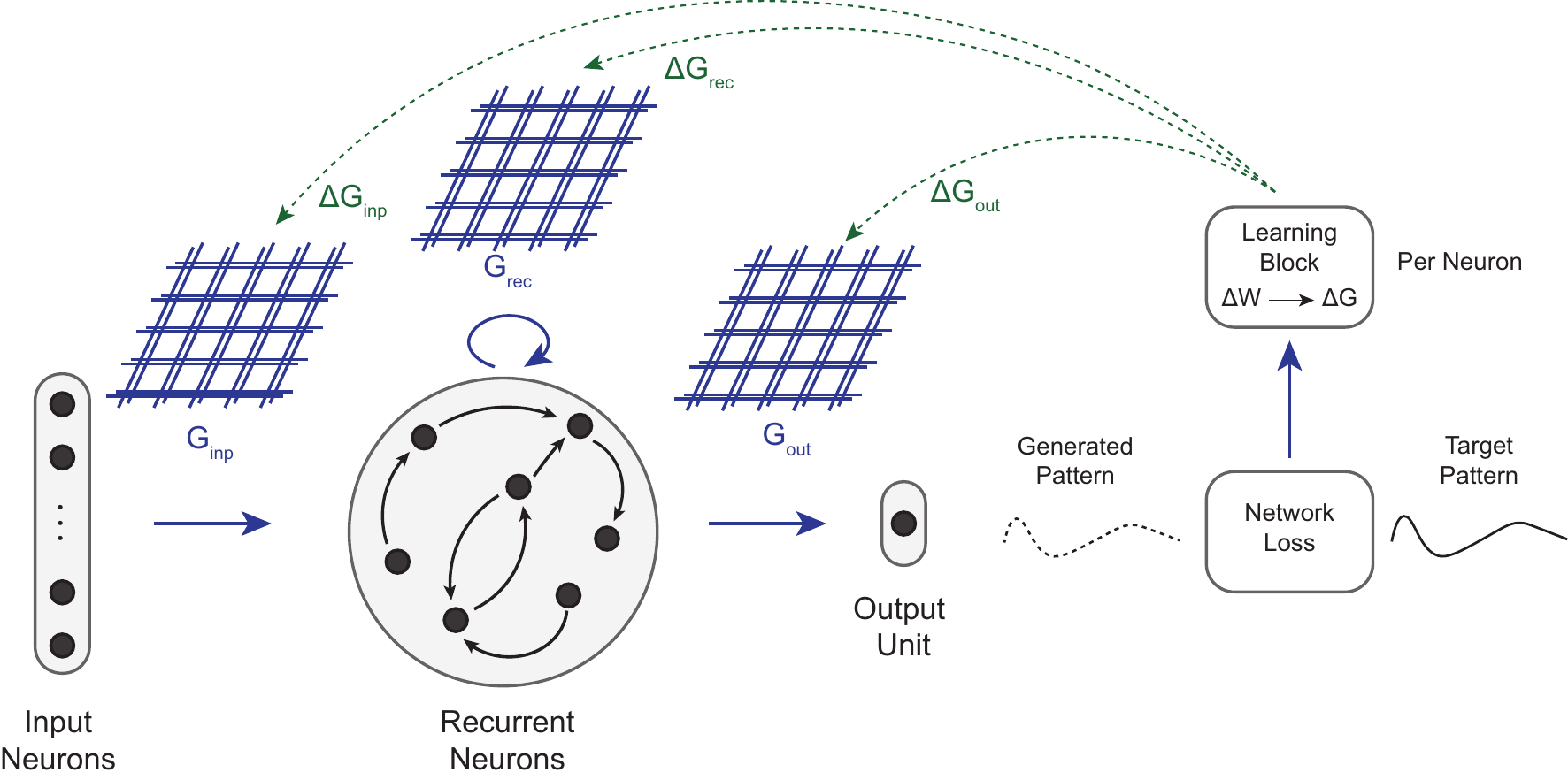}
  \caption{Overview of the spiking \ac{RNN} training framework of the proposed \ac{PCM} crossbar array simulation framework, illustrated here for a pattern generation task.
    The network weights are allocated from three crossbar array models, $G_{inp}$, $G_{rec}$, $G_{out}$.
    The network-generated pattern and the target pattern are compared to produce the learning signal which is fed back to every neuron.
    The \ac{LB} calculates instantaneous weight changes $\Delta W$ using the e-prop learning rule and has to efficiently transfer the desired weight change to a conductance change, i.e. $\Delta W \rightarrow \Delta G$, considering \ac{PCM} non-idealities.}
  \label{fig:eprop-pcm}
\end{figure}

In this work, we simulated the neuron dynamics using the \ac{LIF} neuron model~\cite{Gerstner_Kistler02,Dayan_Abbott01}. 
\ac{LIF} neurons are stateful through their membrane potential, and their temporal dynamics in a spiking \ac{RNN} can be simulated with the following discrete-time equations~\cite{Bellec_etal20}:

\begin{equation}
\begin{aligned}
  v_{j}^{t+1}&=\alpha v_{j}^{t}+\sum_{i \neq j} W_{j i}^{\mathrm{rec}} z_{i}^{t}+\sum_{i} W_{j i}^{\mathrm{in}} x_{i}^{t}-z_{j}^{t} v_{\mathrm{th}} \\
  z_{j}^{t}&=H\left(\frac{v_{j}^{t}-v_{\mathrm{th}}}{v_{\mathrm{th}}}\right)\\
  y_{k}^{t+1} &=\kappa y_{k}^{t}+\sum_{j} W_{kj}^{\mathrm{out}} z_{j}^{t}
\end{aligned}
\label{eq:forward}
\end{equation}

where $v_j^t$ is the membrane voltage of neuron $j$ at time $t$.
The output state of a neuron is a binary variable, $z_j^t$, that can either indicate a spike, $1$, or no spike, $0$.
The neuron spikes when the membrane voltage exceeds the threshold voltage $v_{th}$, a condition that is implemented based on the Heaviside function $H$.
The parameter $\alpha \in [0,1]$ is the membrane decay factor calculated as $\alpha=e^{-\delta t/ \tau_m}$, where $\delta t$ is the discrete time step resolution of the simulation and $\tau_m$ is the neuronal membrane decay time constant, typically tens of milliseconds. 
The network activity is driven by input spikes $x_i^t$.
Input and recurrent weights are represented as $W_{ji}^{\mathrm{in}}$ and $W_{ji}^{\mathrm{rec}}$ respectively.
At the output layer, the recurrent spikes are fed through readout weights $W_{kj}^{\mathrm{out}}$ into a single layer of leaky integrator units $y_{k}$ with the decay factor $\kappa \in [0,1]$.
This continuous valued output unit is analogous to a motor function which generates coherent motor output patterns~\cite{Sussillo_Abbott09} of the type shown in Fig.~\ref{fig:eprop-pcm}.

The aim of the training process is to find optimal network weights $\{W_{ji}^{\mathrm{inp}}, W_{ji}^{\mathrm{rec}}$ and $W_{kj}^{\mathrm{out}}\}$, that maximize the performance of the network~\cite{Richards_etal19}.
One approach is to train the spiking \acp{RNN} offline, using any optimization method, and then transfer the weights to the crossbar via an iterative procedure for inference applications~\cite{Diehl_etal15}.
However this would not take into account the non-idealities of \ac{PCM} synapses.
Our aim is to develop an online training framework that can compensate for these non-idealities.
To be suitable for neuromorphic hardware, this framework must use a learning algorithm that is (i) local, (ii) online and (iii) well-tested beyond toy problems. 
For example, the FORCE algorithm performs well on motor tasks~\cite{Nicola_Clopath17a, Sussillo_Abbott09}, however the weight updates require the knowledge of all synaptic weights, which violates the first requirement.
Similarly, training the network with \ac{BPTT} using surrogate gradients \cite{Neftci_etal19,Lee_etal16} is not an option as it requires buffering all intermediate neuron states to calculate updates backward in time, which violates the second requirement.

One promising solution to the credit assignment problem in the spiking \ac{RNN} landscape is provided by the e-prop algorithm~\cite{Bellec_etal20}.
It has been shown that e-prop can offer an accuracy similar to that of \ac{LSTM} networks \cite{Hochreiter_Schmidhuber97} trained with \ac{BPTT} on complex temporal tasks.
As this algorithm works by factorizing the gradients of \ac{BPTT} as a sum of products between instantaneous learning signals and local eligibility traces, it is both online and local.
Similar to e-prop, the recently published \ac{OSTL} algorithm~\cite{Bohnstingl_etal20} also supports complex recurrent architectures, and in addition it is able to generate the exact gradients calculated with \ac{BPTT}.

\subsection{Memristor-Aware Weight Update Optimization}
\label{section:updates}
Many neuromorphic processors with on-chip learning include a \ac{LB} connected to the neuron circuits~\cite{Qiao_etal15,Payvand_etal18,Payvand_etal19}.
The function of \ac{LB} is to continuously compare the activity of the neuron with a desired target activity, and (ii) estimate the amount of required change in the neuron's synaptic weights, based on the implemented learning rule and error function.
When the weight update condition is met, which can be on the arrival of the pre-synaptic spike, on the post-synaptic spike, or on a error-driven signal~\cite{Payvand_etal20}, the \ac{LB} instructs the corresponding synapses to update the weight with the calculated weight change.
However, memristive synapses cannot be easily programmed to target specific conductances, due to device non-idealities such as limited bit precision, asymmetric SET/RESET updates and programming noise. 

Weight update mechanisms in memristive architectures are primarily designed to cope with memristive non-idealities to execute the desired weight changes on synapses.
These mechanisms are typically single-shot, i.e. one or multiple gradual SET pulses are applied without reading the device state.
This avoids iterative write-read-verify schemes or modifications of the pulse shape during the training, and hence enables simpler circuits with a better energy efficiency.
In our framework we use a differential synaptic configuration~\cite{Nandakumar_etal18, Boybat_etal18, Nandakumar_etal20a}, where every synapse keeps two sets of memristors ($G^+$ and $G^-$) whose difference gives the effective synaptic conductance (Fig.~\ref{fig:mushroom}a), so that both positive and negative synaptic weights can be represented using memristor conductances.

With this differential configurations however the unidirectional updates used to update the synaptic weight may result in the saturation of either or both of the devices  in the synapse~\cite{Nandakumar_etal18, Boybat_etal18, Nandakumar_etal20a}.
One solution to the problem would be to employ a push-pull mechanism: to increase the synaptic efficacy, the positive memristor conductance is increased while the negative memristor conductance is decreased~\cite{Nair_etal17}.
Unfortunately this mechanism is not compatible with \ac{PCM} devices as the melt-quenching-based RESET is an abrupt process~\cite{Burr_etal14}, thus requiring a refresh mechanism by resetting both positive and negative memristors and reprogramming them to the their effective conductances when specific criteria is met.
In the following, we will describe four weight update methods that the neuromorphic processors have widely adopted and are implemented in the PCM crossbar array simulation framework.

\paragraph{The Sign Gradient Descent.}
In this method, synaptic weights are updated according to the sign of the gradient of the loss function, estimated by the learning algorithm.
The idea of the sign gradient descent is to take a fixed step ($\delta$) in the direction of descending gradient, but neglecting its magnitude.
Nevertheless, under some assumptions, the convergence is guaranteed~\cite{Balles_etal20a}. The synaptic weight updates $\Delta W$ can be computed as
  
\begin{equation}
	\Delta W = -\delta \operatorname{sign}(\nabla \mathcal{L}) SL,
  \label{eq:sgd}
\end{equation}

where $\delta$ is the amount of change, $\nabla \mathcal{L}$ is the gradient calculated by the chosen learning rule, and $SL = |\nabla{\mathcal{L}}|>\theta$ is a binary variable indicating a stop-learning regime. The latter increases the stability during learning, which enables updates only when the magnitude of the gradient is higher than a fixed threshold $\theta$.

Due to its simplicity, the sign gradient descent is popular among memristive neuromorphic systems~\cite{Nair_Dudek15, Muller_etal17, Payvand_etal20a}.
When implementing it with memristor synapses, the \ac{LB} sends a single UP (or DOWN) pulse to instruct an increase (or decrease) of the synaptic weights.
Hence, on the onset of the weight update, a single SET pulse is applied either to the $G^+$ or the $G^-$ \ac{PCM} device, determined by the sign of the gradient.
However, the effective value of $\delta$ is not a constant due to the WRITE noise, and not symmetric because SET operation in \ac{PCM} is gradual whereas RESET is abrupt.

\paragraph{Stochastic Update.}
Conventional optimization methods demand the amount of weight change at every
weight update to be 3-4 orders of magnitude smaller than the original weight~\cite{Kingma_Ba14}, usually requiring at least an 8-bit weight representation, whereas a typical \ac{PCM} device can only represent 3-4 bits of information~\cite{Athmanathan_etal16a}.

Therefore, even a single WRITE pulse applied to a \ac{PCM} device leads to an order-of-magnitude overshoot compared to the desired amount of weight change typically calculated by the \ac{LB}.
One method used by Nandakumar et al.~\cite{Nandakumar_etal18} aims at overcoming this scaling disparity by stochastically executing (or skipping) the weight updates depending on the magnitude of the gradient.

We implemented the stochastic weight update on our \ac{PCM} crossbar simulation by scaling the gradient magnitude with a scaling factor, $p$, to represent the update probability such that
\begin{equation}
P(\text{update}) = \frac{|\nabla \mathcal{L}|}{p}
\end{equation}
By tuning the scaling factor, the number of devices that get programmed can be controlled. 
Single pulses are applied accordingly to the synapses.
The refresh criteria, unlike the original work~\cite{Nandakumar_etal18}, is checked and applied before the weight update to prevent the execution of the update on saturated devices.

\paragraph{Multi-memristor Update.}
To increase the dynamic range and the resolution of the synaptic weights, the concept of multiple \ac{PCM} devices per synapses has been proposed in~\cite{Boybat_etal18}, where each synapse consists of $2N$ memristors ($N$ positive and negative pairs in a differential configuration), the total conductance determining the synaptic efficacy.
The updates are applied to the devices sequentially, which in turn decreases the smallest mean weight change by a factor of $2N$ and reduces the variance due to WRITE noise by a factor of $\sqrt{2N}$~\cite{Boybat_etal18} (see \nameref{section:transfer}).

Our multi-memristor implementation first estimates the the number of pulses required to match the desired conductance change, by assuming each SET pulse increases the conductance linearly by $0.75$ $\mu$S as the model outlined in Section~\ref{section:model} suggests (a more realistic approximation of the number of pulses is available in~\cite{Nandakumar_etal18}).
Then these SET pulses are applied sequentially among $N$ \ac{PCM} devices in a circular queue.
The refresh is applied if any of the memristor pair conductances is higher than 9 $\mu$S and their difference is less than $4.5\mu$S.

\paragraph{Mixed-precision Update.}
One solution to close the gap between the update resolution requested by the \ac{LB} and the minimum programmable conductance change of \acp{PCM} is to accumulate the updates on a high-precision co-processor until they become reliably transferable to \acp{PCM}~\cite{Nandakumar_etal20a}.
Notably, this scheme also corresponds to the quantization-aware training techniques conventionally used for the training of quantized artificial neural networks~\cite{Hubara_etal16}.

In our simulations, the \ac{LB} accumulates the gradients in a double-precision floating memory until they are an integer multiple of \ac{PCM} update granularity (corresponding to $0.75$ $\mu$S).
Then they are converted to a number of pulses and applied to the memristors.
The refresh is applied if either one of the memristor pair conductance is higher than 9 $\mu$S and their difference is less than $4.5\mu$S.

\section{Training a Spiking RNN on a \ac{PCM} Crossbar Simulation Framework}
\label{section:trainingsnn}

\FloatBarrier

We used the model of the \ac{PCM} crossbar array to determine realistic values of the network parameters $\{W_{ji}^{\mathrm{in}}$, $W_{ji}^{\mathrm{rec}}$ and $W_{kj}^{\mathrm{out}}\}$.
In order to represent synaptic weights, $W \in [-1, 1]$ , with the conductance values of \ac{PCM} devices, $G \in [0.1, 12]$ $\mu$S~\cite{Nandakumar_etal18}, we used the linear relationship $W = \beta [\sum_{N} G^+ - \sum_{N} G^-]$, where $\sum_N G^+$ and $\sum_N G^-$ are the total conductance of $N$ memristors\footnote{$N=1$ for all weight update methods, except when the multi-memristor update is being used.} representing the potentiation and the depression of the synapse respectively~\cite{Nandakumar_etal18}.
At this stage, the forward computation (inference) of Eq.~\ref{eq:forward} is simulated using the \ac{PCM} crossbar simulation framework that includes the effects of READ noise and temporal conductance drift.

The weight updates calculated by the e-prop algorithm are applied to the \ac{PCM}-based crossbar arrays using each of the methods described in Section~\ref{section:updates}.

\section{Results}
\label{section:results}

We validated the network using a classical one-dimensional continuous pattern generation task relevant for many domains including motor control or value function estimation in \ac{RL}\cite{Sussillo_Abbott09}
The training dataset is the slightly modified version of the one-second-long patterns described in \cite{Bellec_etal20} (see \nameref{section:methods} for task and training details).

\begin{figure}[!ht]
  \centerfloat
  \includegraphics[width=1.1\textwidth]{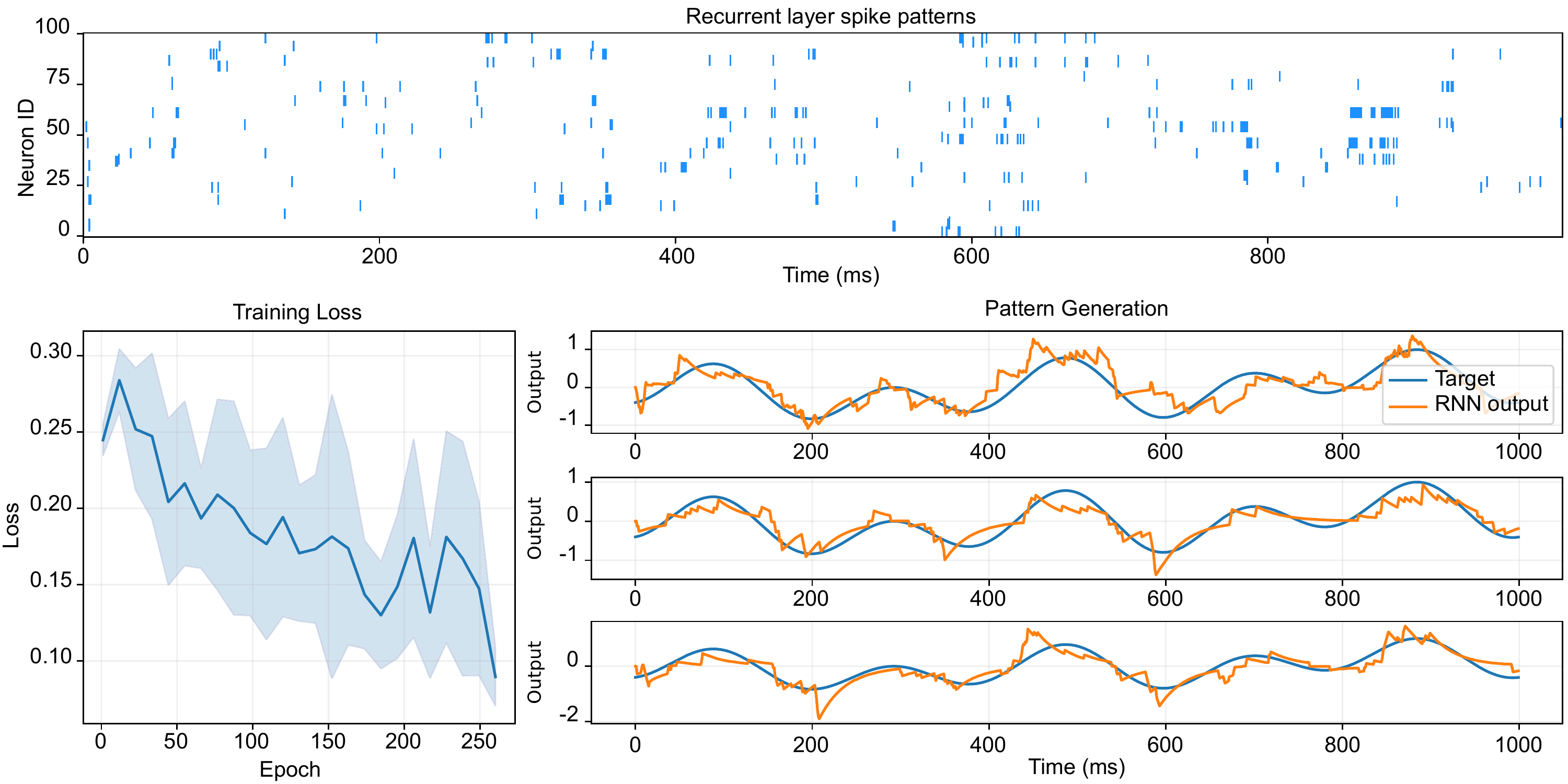}
  \caption{Dynamics of a network trained with the mixed-precision algorithm. The raster plot (top) shows the sparse spiking activity of recurrent-layer neurons. The training loss (bottom left) demonstrates MSE loss over 250 epochs is averaged over ten best network hyperparameters (see Fig~\ref{fig:losstr} for the best performing hyperparameter). Properly-tuned neuronal time constants and trained network weights result in generated patterns following the targets (bottom right). The generated patterns are extracted from three different spiking \acp{RNN}.}
  \label{fig:trainingstats}
\end{figure}

\begin{figure}[!ht]
  \centering
  \includegraphics[width=0.85\textwidth]{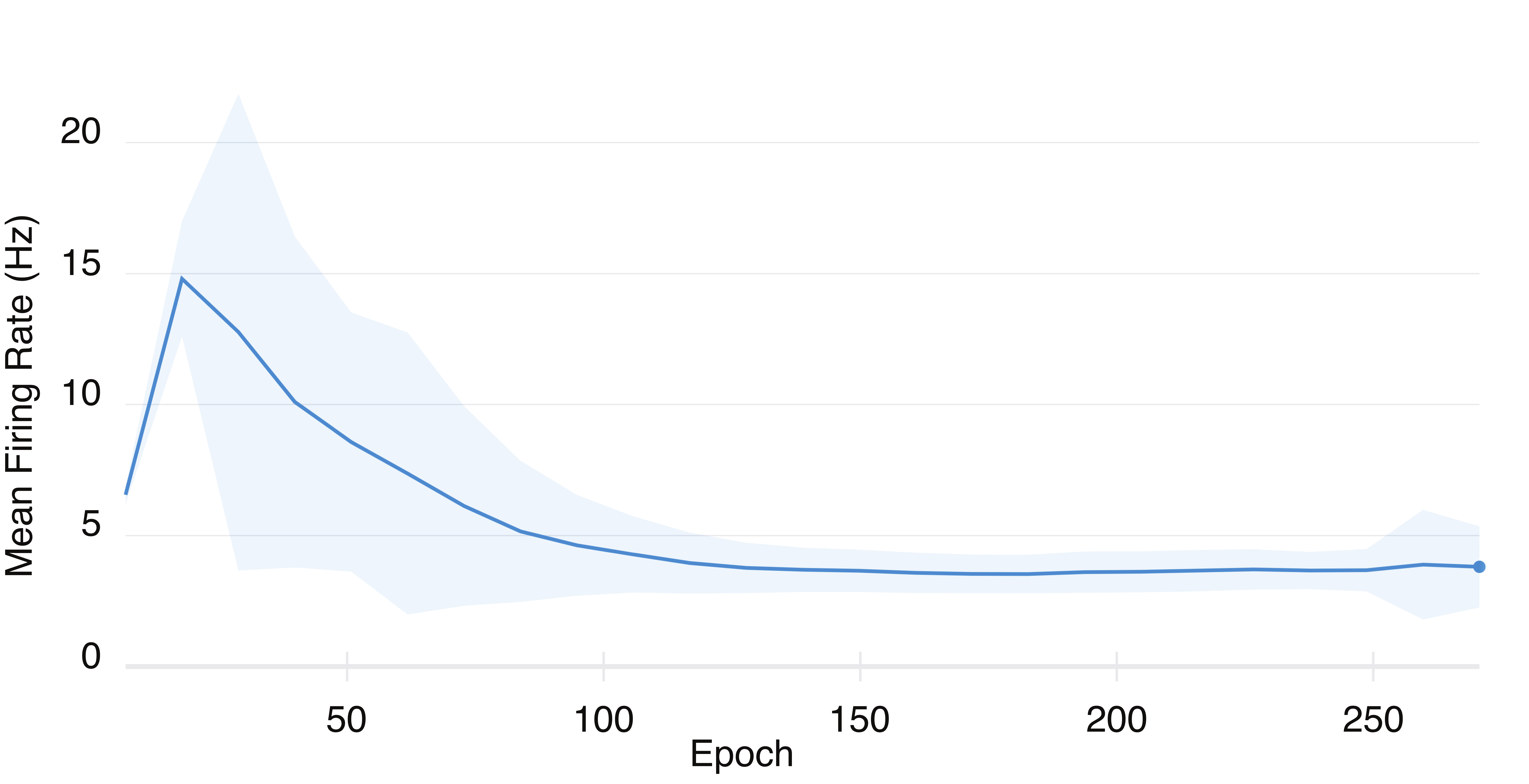}
  \caption{Mean firing rate of 50 networks with \ac{PCM} synapses trained using the mixed-precision method.}
  \label{fig:fr}
\end{figure}

Table~\ref{comparison-table-xbar} summarizes the training performances of spiking \acp{RNN} utilizing different weight update methods on \ac{PCM} crossbar arrays to realize the target weight change calculated by the e-prop algorithm.
Out of five configurations, only the mixed-precision approach resulted in an acceptable performance on the pattern generation task (MSE loss < 0.1, see Section~\ref{section:loss-eval} for the evaluation). 
Figure~\ref{fig:trainingstats} demonstrates extremely sparse spiking activity of about $3.3$ Hz in the recurrent layer (see Fig.~\ref{fig:fr} for mean firing rate evolution during the training), nevertheless the network is able to generate the target patterns well despite \ac{PCM} non-idealities.

\begin{table}
  \caption{Performance evaluation of spiking \acp{RNN} with models of \ac{PCM} crossbar arrays.}
  \label{comparison-table-xbar}
  \centerfloat
  \begin{tabular}{lllllll}
    \toprule
    \textbf{Method}   & Sign-gradient & Stochastic &  Multi-mem (N=4) & Multi-mem (N=8) & Mixed-precision\\
    \midrule
    \textbf{MSE Loss} & 0.2080        & 0.1808     & 0.1875           &  0.1645         &  0.0380 \\
    \bottomrule
  \end{tabular}
\end{table}

\FloatBarrier

\begin{figure}
  \centerfloat
  \includegraphics[width=1.1\textwidth]{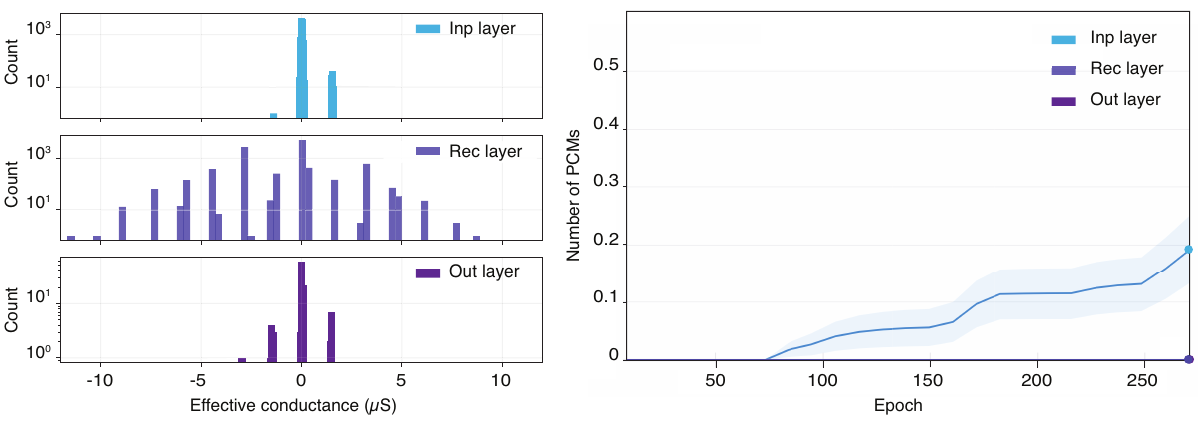}
  \caption{(left) The effective conductance distributions $(G^+ - G^-)$ of the synapses in the input, recurrent and output layer, at the end of the training with the mixed-precision method. (right) Averaged over 50 training runs, the mean number of \ac{PCM} devices requiring a refresh is shown for each layer. The refresh operation was not needed for recurrent and output layers.}
  \label{fig:devicestats}
\end{figure}
\FloatBarrier

\begin{figure}[!ht]
\centerfloat
\includegraphics[height=3.0cm, keepaspectratio]{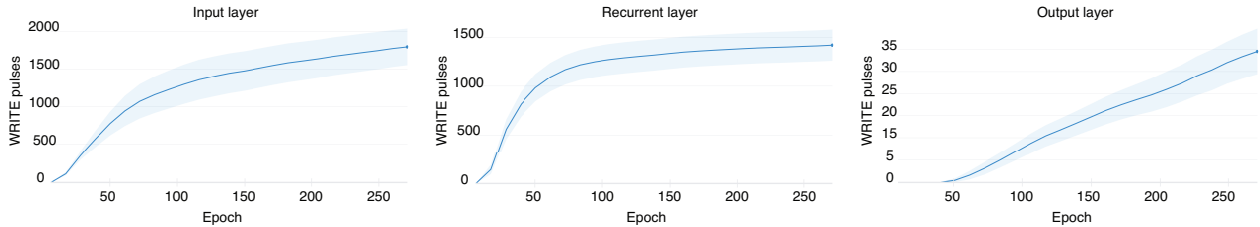}
\caption{Total number of WRITE pulses applied to \ac{PCM} devices are shown for the input, recurrent and output layers. Only 0.07\%, 0.07\% and 0.1\% of \ac{PCM} devices within each layer are programmed respectively during mixed-precision training.}
\label{fig:pulses}
\end{figure}

We also observed that, with our training settings, the weight saturation problem due to the differential configuration does not occur often and only a few devices ($<1\%$, as shown in Fig~\ref{fig:devicestats} (right)) are required to be refreshed.
This is partially because the total number of WRITE pulses applied during the training is very low, i.e. only $\sim 12$ WRITE pulses are applied per epoch, as shown in Fig.~\ref{fig:pulses}.
Thanks to the mixed-precision algorithm, only large-enough accumulated gradients lead to the generation of WRITE pulses.
Fig.~\ref{fig:devicestats} (left) demonstrates the effective weight distribution of the \ac{PCM} synapses at the end of the training.

To simulate an ideal device model, we disabled all noise and drift effects in the simulation framework (see \nameref{section:xbarperf}). 
We kept a limited weight resolution of 4-bit, an optimistic but achievable target for \ac{PCM} devices~\cite{Gallo_etal18}.
This ideal \ac{PCM} model is therefore equivalent to a digital 4-bit memory.
Table~\ref{comparison-table-perf} summarizes the performances of networks utilizing different memristor-aware weight update methods with this ideal model.
The results show that the stochastic updates, the multi-memristor updates with $N=8$, and the mixed-precision updates can solve the pattern generation task. 
Similarly to the realistic device model results in Table~\ref{comparison-table-xbar}, the mixed-precision method achieved the best accuracy. 
Unsurprisingly, all methods performed better in the absence of the previously discussed PCM non-idealities. 
One remark is that stochastic updates allow for a better performance than both multi-memristor methods, indicating the necessity of having a low number of stochastic updates when the network is trained with the quantized weights.

Moreover, to further evaluate the effect of limited bit precision on the network performance,  we trained the same network with e-prop using standard FP32 (single-precision floating point) weights. 
With the the high resolution of the FP32 training, we achieved an mean squared error (MSE) loss of 0.0215, which is comparable to mixed-precision training using either ideal quantized memory or \ac{PCM} cell model.

\begin{table}[!ht]
   \caption[Performance evaluation of spiking \acp{RNN} with an ideal crossbar array model]{Performance evaluation of spiking \acp{RNN} with an ideal crossbar array model\protect\footnotemark}

  \label{comparison-table-perf}
  \centerfloat
  \begin{tabular}{lllllll}
    \toprule
    \textbf{Method}   & Sign-gradient & Stochastic &  Multi-mem (N=4) & Multi-mem (N=8) & Mixed-precision\\
    \midrule
    \textbf{MSE Loss} & 0.1021        & 0.0758     &  0.1248          &  0.0850         &  0.0289 \\
    \bottomrule
  \end{tabular}
\end{table}
\footnotetext{Multi-memristor configurations are implemented assuming a 4-bit resolution per memory cell. Hence $N=4$ and $N=8$ is equal to having 7-bit and 8-bit resolutions digital weights per synapse, respectively.}

\begin{figure}[!ht]
  \centering
  \includegraphics[width=0.9\textwidth]{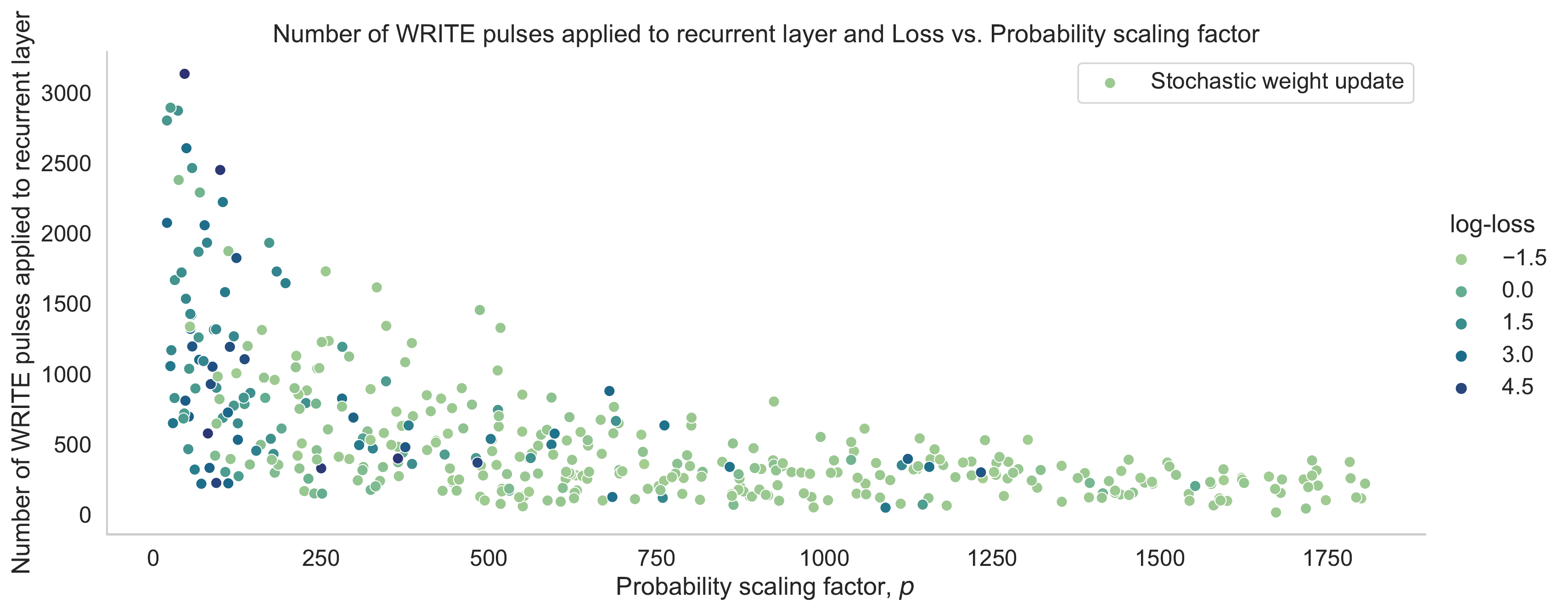}
  \caption{The stochastic update method enables tuning the number of WRITE pulses to be applied to \ac{PCM} devices.}
  \label{fig:stoch}
\end{figure}

Similar to \cite{Nandakumar_etal18}, we observed that probability scaling factor $p$ in the stochastic update method allows the tuning of how many number of devices are being programmed during the training. 
Figure~\ref{fig:stoch} demonstrates that with the increase of $p$ (decreasing update probability), the number of WRITE pulses applied to \ac{PCM} devices can be decreased up to one order of magnitude without degrading the loss.
  
\FloatBarrier

\section{Discussion}

On-chip training of spiking \acp{RNN} enables the low-power deployment of the intelligent computing systems at the edge with learning and adaptation capabilities~\cite{davies_etal21,frenkel_etal21}.
In this work, we evaluated four widely used memristor update mechanisms on spiking \acp{RNN} based on the ideal gradient information calculated by the e-prop learning rule. 
Crossbar architectures with memristive devices intrinsically support the sparse, event-driven, and asynchronous processing of spike flow through in-memory computing, reducing the dynamic power consumption and shrinking the memory area.
Implementing learning on-chip imposes both space and time locality constraints in recurrent architectures, which can be met by the e-prop learning rule, for example. 
By doing extensive hyperparameter exploration, we investigated the performances of weight update mechanisms despite various \ac{PCM} non-idealities on a second-long pattern-generation task. Solving the pattern generation task demonstrates the recurrent network's ability to preserve task-relevant information in its activation dynamics for at least one second period. Such ability is required for applications processing temporal inputs such as keyword detection, motor control, and bio-signal processing.

Among the mechanisms we studied, the mixed-precision update leads to the best accuracy. 
This is because accumulating instantaneous gradients on high-precision enables the use of a low learning rate, making the ideal weight update magnitude comparable to minimum programmable \ac{PCM} conductance change, resulting in an improved convergence of the network.
Mixed-precision hardware resembles the concept of the cascade memory model in neuroscience, where complex synapses hold a hidden state variable which only becomes visible after hitting a threshold~\cite{Fusi_Abbott07}.
This \textit{meta-plastic} model has recently been used for solving catastrophic forgetting in some benchmarks~\cite{laborieux_etal21}.
However, the mixed-precision scheme comes at the cost of having a high-precision memory that accumulates the updates.
This has previously been done by introducing a co-processor next to the crossbar memristor array~\cite{Nandakumar_etal20a}.
Yet, accumulating the gradients allows reliable \ac{PCM} programming, reducing the number of programming devices and speeds up the learning.
Therefore the synergy between memristor-based synapses and learning rules that intrinsically allows the accumulation of gradients, e.g., e-prop, is remarkable and worth exploring.

One of the key levers to increase the learning performance is the memory resolution, which comes in line with the mixed-precision learning study of \cite{Nandakumar_etal20a}.
However, this comes at the price of increasing the synapse size by increasing the number of devices.
One solution to this problem is to employ binary synapses with a stochastic rounding update scheme~\cite{Muller_Indiveri15a}.
Introducing stochasticity reduces the learning rate and, in expectation, moves the weight parameters to the optimal binary value. Stochastic rounding approaches have already been employed successfully not only in memristive devices but also in fully-digital designs to apply low learning rate values to quantized weights~\cite{Frenkel_etal19, Frenkel_etal20}.
It has also been shown that the intrinsic cycle-to-cycle variability of memristive devices can implement stochastic updates in an area-efficient manner~\cite{Payvand_etal19}.

To the best of our knowledge, there is no report yet on the online training of spiking \ac{RNN} with e-prop learning rule based on realistic \ac{PCM} synapse models.
Our work compares several previously developed methods that are designed to cope with memristor non-idealities and demonstrates that accumulating gradients allows more reliable programming of \ac{PCM} devices, reduces the number of programming devices and outperforms other synaptic weight-update mechanisms.
Future work will need to evaluate the impacts of the implemented weight update schemes using more extensive datasets for a more interpretable benchmarking, further incorporate the \ac{PCM} devices for emulating temporal dynamics such as eligibility traces~\cite{Demirag_etal21a}, as well as exploring other learning rules such as \ac{OSTL} or \ac{RTRL}~\cite{Williams_Zipser89}.


\section{Methods}
\label{section:methods}
For the chosen pattern generation task, the network consists of 100 input and 100 recurrent \ac{LIF} neurons, and one leaky-integrator output unit.
The network receives a fixed-rate Poisson input, and the target pattern is a one-second-long sequence defined as the sum of four sinusoids (1\,Hz, 2\,Hz, 3\,Hz and 5\,Hz), whose phases and amplitudes are randomly sampled from uniform distributions $[0, 2\pi]$ and $[0.5, 2]$, respectively.
Throughout the training, all layer weights \{$W_{ji}^{\mathrm{in}}$, $W_{ji}^{\mathrm{rec}}$ and $W_{kj}^{\mathrm{out}}$\}, are kept plastic and the device conductances are clipped between 0.1 and 12 $\mu$S. This benchmark is adapted from~\cite{Bellec_etal18}.

We trained $\sim 1000$ different spiking \acp{RNN} for each of the weight update methods described in Section \ref{section:updates}. 
Each network shares the same architecture, except for their synapse implementations, some of their hyperparameters and weight initialization.
Because each weight update method requires a few specific additional hyperparameters and may considerably affect the network dynamics, we tuned the network hyperparameters for each of the update methods using Bayesian optimization~\cite{Snoek_etal12a}. . 
Based on network performances over 250 epochs of the pattern generation task, we selected the best performing network hyperparameters out of 1000 candidates.
By doing so, we evaluated how well different weight update methods can reflect the ideal weight update calculated by the e-prop learning rule on a \ac{PCM} substrate.


\section{Acknowledgements}
This project has received funding from the European Union's H2020 research and innovation programme under the Marie Skłodowska-Curie grant agreement No 861153, Memscales project (871371), and ERA-NET CHIST-ERA programme by SNSF (20CH21186999 / 1).
\newpage
\bibliographystyle{unsrt}
\bibliography{main}

\newpage
\section{Supplementary}
\subsection*{Supplementary Note 1}
\label{section:xbar}

We implemented the \ac{PCM} crossbar array simulation framework in PyTorch~\cite{Paszke_etal19a}, which can be used for both the inference and the training of \acp{ANN} or \acp{SNN}.
Built on top of the statistical model introduced by Nandakumar et al.~\cite{Nandakumar_etal18}, our crossbar model supports asynchronous SET, RESET and READ operations over entire crossbar structures and simultaneously keep tracks of the temporal evolution of device conductances.

A single crossbar array consists of $P \times Q$ nodes (each node representing a synapse), where single node has $2N$ memristors arranged using the differential architecture ($N$ potentiation, $N$ depression devices).
Each memristor state is represented by four variables, $t_p$ for storing the last time the device is written (which is used to calculate the effect of the drift), $count$ for counting how many times it has written (to be used later in the arbiter of N-memristor architectures), $P_{mem}$ for its programming history (required by the \ac{PCM} model) and $G$ for representing the conductance of the the device at $T_0$ seconds later after the last programming time.
The initial conductances of \ac{PCM} devices in the crossbar array are assumed to be iterativelly programmed to \ac{HRS}, sampled from a Normal distribution $\mathcal{N}(\mu=0.1,\sigma=0.01)$ $\mu$S.

The \ac{PCM} crossbar simulation framework supports three major functions: READ, SET and RESET.
The READ function takes the pulse time of the applied READ pulse, $t$, and calculates the effect of drift based on the last programming time $t_p$. 
Then, it adds the conductance-dependent READ noise and returns the conductance values of whole array.
The SET function takes the timing information of the applied SET pulse, together with a mask of shape $(2 \times N \times P \times Q)$ and calculates the effect of the application of a single SET pulse on the \ac{PCM} devices that are selected with the mask.
Finally, the RESET function initializes all the state variables of devices selected with a mask and initializes the conductances using a Normal distribution $\mathcal{N}(\mu=0.1,\sigma=0.01)$ $\mu$S.

\subsection*{Supplementary Note 2}
\label{section:xbarperf}
READ and WRITE operations to simulated \ac{PCM} devices in the
crossbar model are stochastic and subject to the temporal conductance drift.
Additionally, \ac{PCM} devices offer a very limited bit precision.
Therefore, to ease the network training procedure, especially the hyperparameter
tuning, we developed the \texttt{perf-mode}.
When crossbar model is operated in \texttt{perf-mode}, all stochasticity sources
and the conductance drift are disabled. READ operations directly access
the device conductance without $1/f$ noise and drift, whereas SET
operations increase the device conductance as

\begin{equation}
  G_{N} = G_{N-1} + \frac{G_{MAX}}{2^{CB_{RES}}},
\end{equation}

where, $G_{MAX}$ is the maximum \ac{PCM} conductance set to 12 $\mu$S (conductivity boundaries are determined based on the device measurements  from \cite{Nandakumar_etal18}), and
$CB_{RES}$ is the desired bit-resolution of a single \ac{PCM} device. 
In a nutshell, the \texttt{perf-mode} turns \acp{PCM} into an ideal memory cells corresponding to a digital memory with a limited bit precision.

\begin{figure}[!ht]
  \centering
  \begin{subfigure}{0.5\textwidth}
    \centering
    \includegraphics[width=0.9\linewidth]{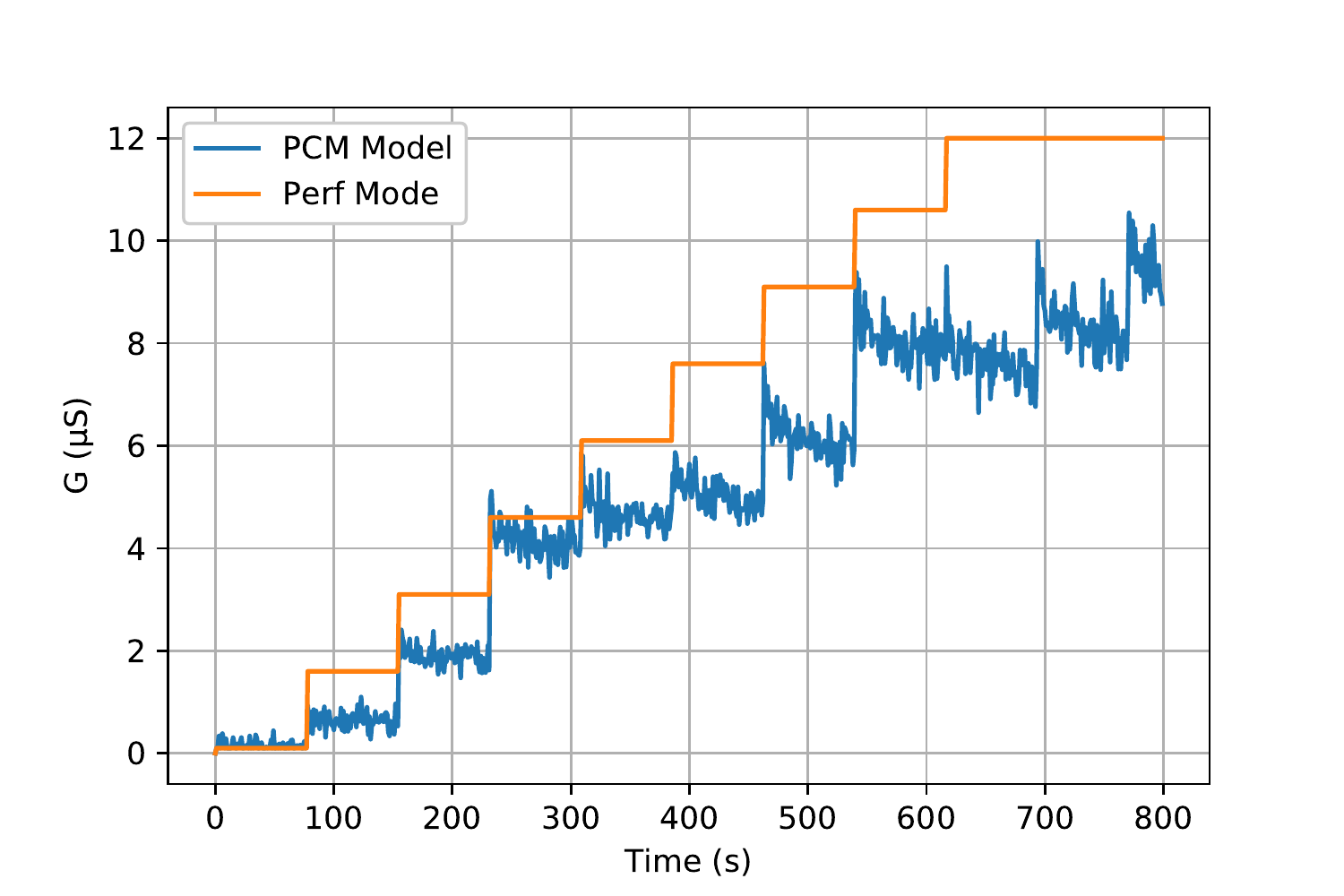}
    \caption{Comparison of the full \ac{PCM} model and its\newline \texttt{perf-mode} equivalent after 8  consecutive SET\newline pulses.}
    \label{fig:sub1}
  \end{subfigure}%
  \begin{subfigure}{0.5\textwidth}
    \centering
    \includegraphics[width=0.9\linewidth]{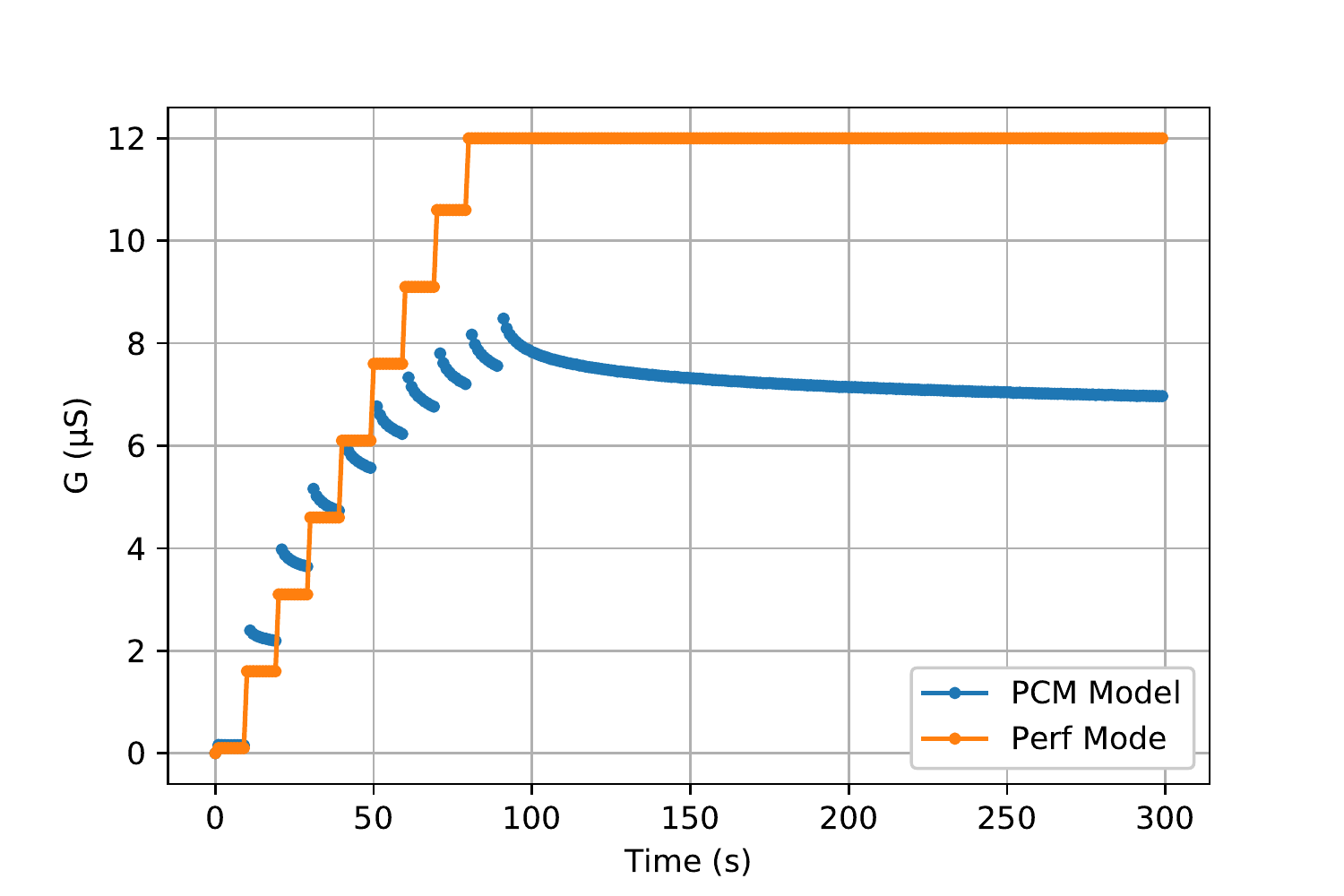}
    \caption{Comparison of the full \ac{PCM} model and its\newline \texttt{perf-mode} equivalent after 8  consecutive SET\newline pulses, averaged over 300 measurements showing \newline the effect of drift.}
    \label{fig:sub2}
  \end{subfigure}
  \caption{The \ac{PCM} crossbar model supports both the full \ac{PCM} model from~\cite{Nandakumar_etal18} and its corresponding simplified version as an ideal digital memory in \texttt{perf-mode}.}
  \label{fig:test}
\end{figure}
\FloatBarrier

\subsection*{Supplementary Note 3}
\label{section:transfer}
Here, we demonstrate the impact of using multiple memristor devices per synapse
(arranged in differential configuration) on the precision of targeted
programming updates.
Specifically, we modeled synapses with $N={1,4,8}$ \ac{PCM} devices and
programmed them from initial conditions of integer conductance values
$G_{source} \in \{-10,10\}$ $\mu$ S  to integer conductance values $G_{target} \in
\{-10,10\}$ $\mu$ S using the multi-memristor update scheme described in Section~\ref{section:updates}. 
The effective conductance of a synapse is calculated by $G_{syn} = \sum_{i=0}^N G_i^+ - \sum_{i=0}^N G_i^-$, however we normalized the
conductance across 1-PCM, 4-PCM and 8-PCM architectures for an easier comparison,
such that $G_{syn} = \frac{1}{N}(\sum_{i=0}^N G_i^+ - \sum_{i=0}^N G_i^-)$.

Our empirical results verifies the claim of Boybat et al. \cite{Boybat_etal18} that the standard deviation and the update resolution of the write process decreases by $\sqrt{N}$.  

\label{section:wtransfer}
\begin{figure}
  \centerfloat
  \includegraphics[width=1.9\textwidth]{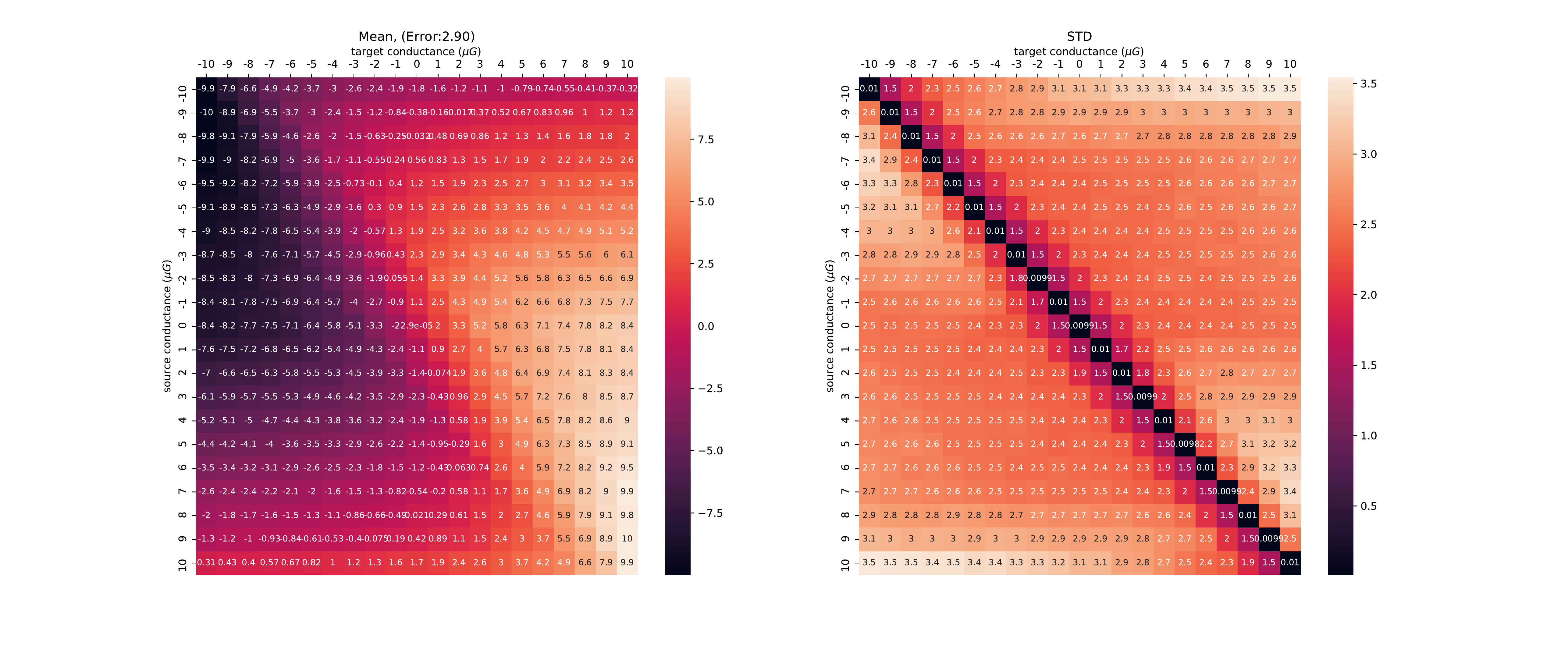}
  \caption{Multi-memristor configuration with 1 \ac{PCM} (one depression and one
    potentiation) per synapse}
  \label{fig:multi1}
\end{figure}
\FloatBarrier

\begin{figure}
  \centerfloat
  \includegraphics[width=1.9\textwidth]{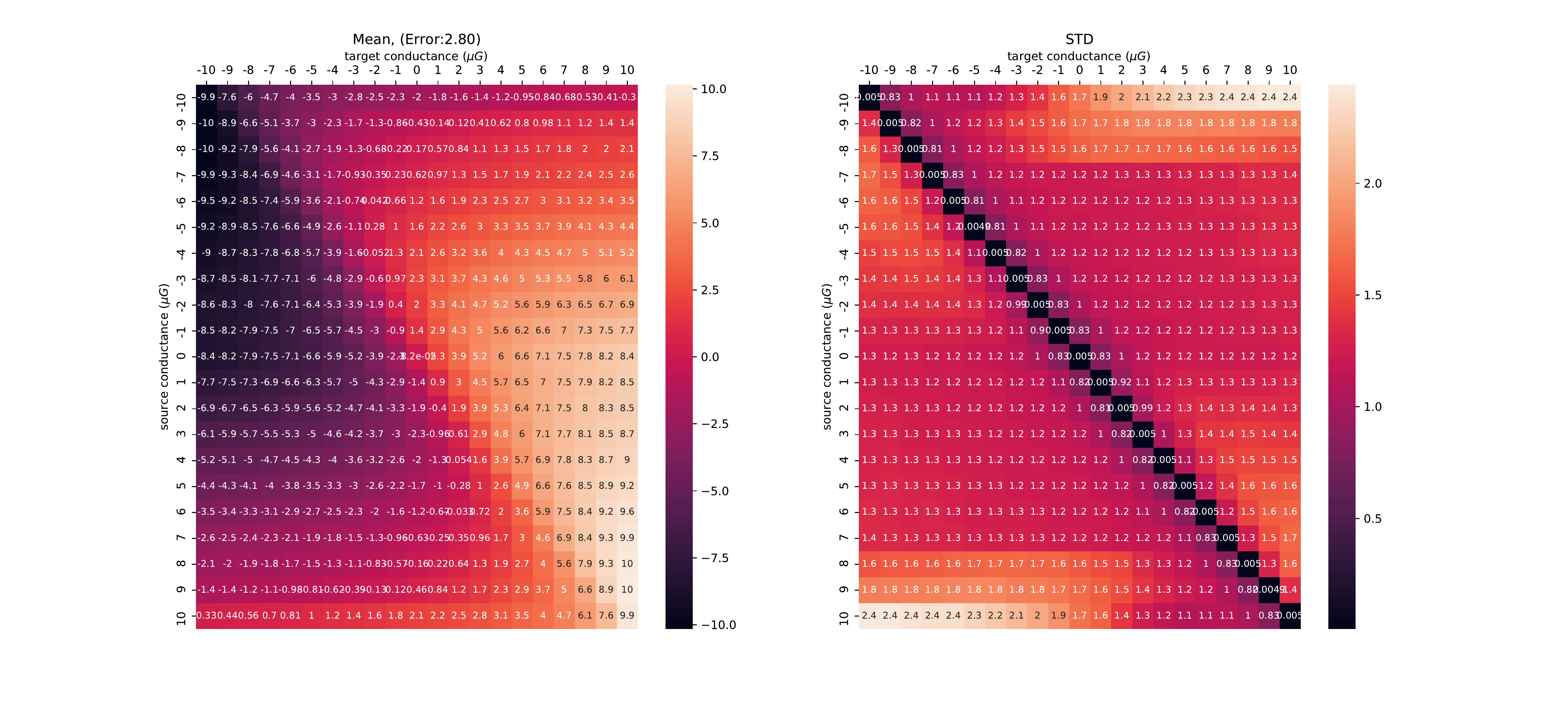}
  \caption{Multi-memristor configuration with 8 \ac{PCM} (four depression and four
    potentiation) per synapse}
  \label{fig:multi4}
\end{figure}
\FloatBarrier

\begin{figure}
  \centerfloat
  \includegraphics[width=1.9\textwidth]{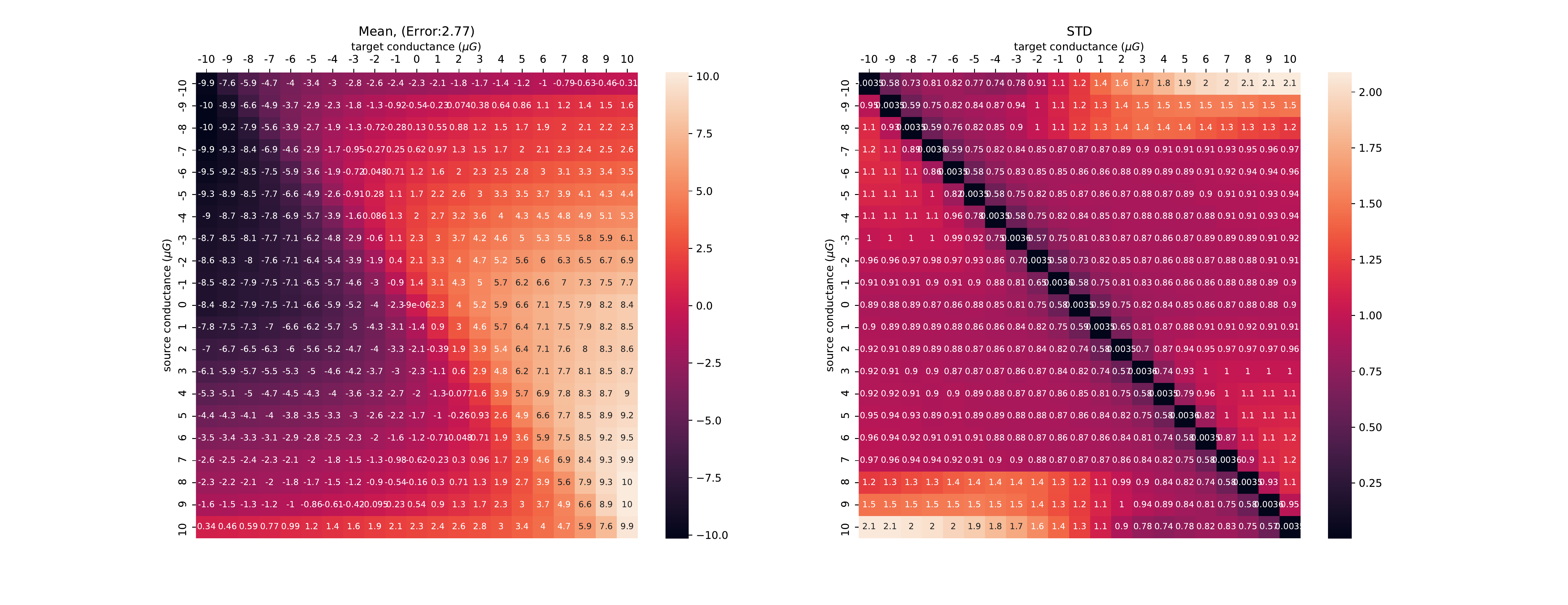}
  \caption{Multi-memristor configuration with 16 \ac{PCM} (eight depression and eight
    potentiation) per synapse}
  \label{fig:multi8}
\end{figure}
\FloatBarrier

\subsection*{Supplementary Note 4}
\label{section:upd-ready}
In the differential architectures, consecutive SET pulses applied to positive and negative memristors may cause the saturation of the synaptic conductance and block further updates.
The saturation effect is more apparent when a single synapse gets 10+ updates in one direction (potentiation or depression) during the training.
For example, this effect is clearly visible in Fig.~\ref{fig:multi1}, Fig.~\ref{fig:multi4} and Fig.~\ref{fig:multi8}, when the source conductance and target conductances differ by more than 8-10 $\mu$S.

We implemented a weight update scheme denoted as the update-ready criterion, which aims to prevent conductance saturation while applying single large updates.
Before doing the update, we read both positive and negative pair conductances, and check if the target update is possible. 
If not, we reset both devices, calculate the new target and apply the number of pulses accordingly.
For example, given $G^+=8 \mu$S and $G^-=4 \mu$S and the targeted update $+6 \mu$S, the algorithm decides to reset both devices because $G^+$ can't be increased up to $14 \mu$S. After both devices are reset, $G^+$ can be programmed to $10 \mu$S).
Although our \ac{PCM} crossbar array simulation framework supports it, this weight transfer criterion is not used in our simulations because it requires reading the device states during the update.

\begin{figure}[!ht]
  \centerfloat
  \includegraphics[width=1.9\textwidth]{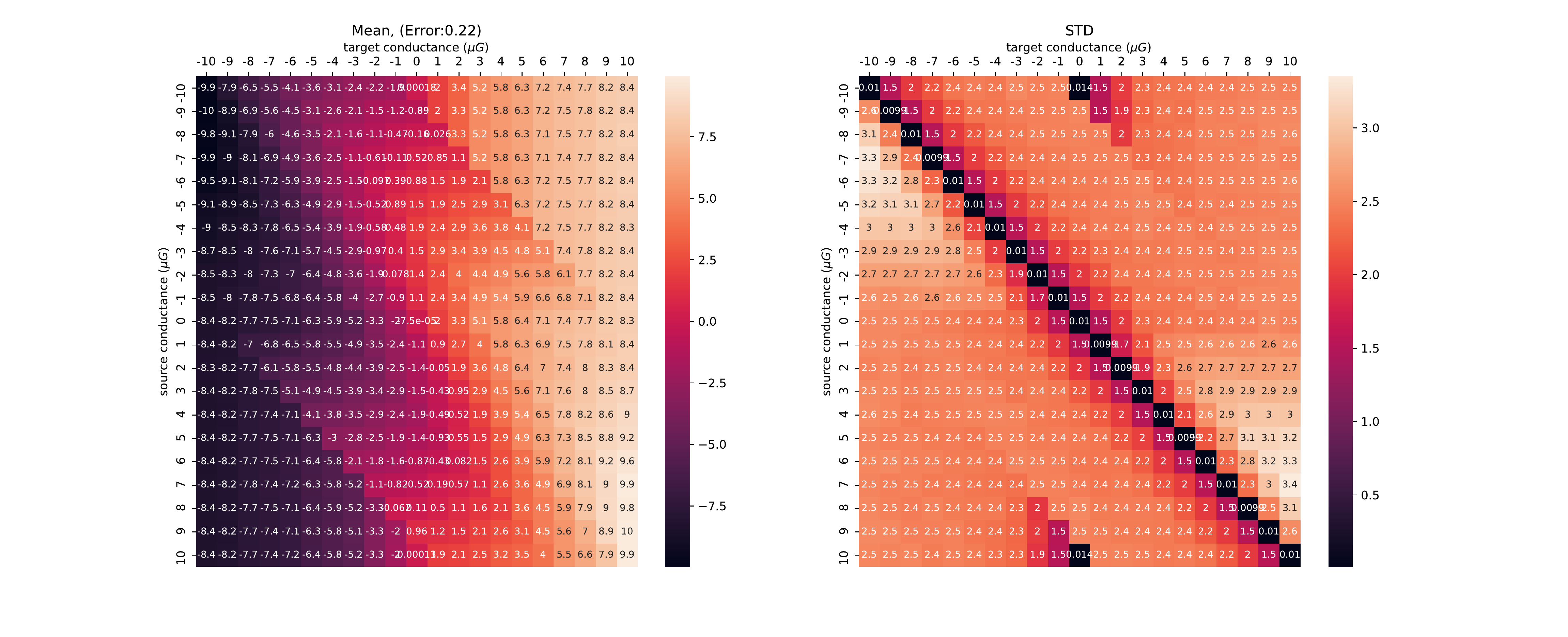}
  \caption{Update-ready criterion tested with $N=1$ memristor per
    synapse.}
  \label{fig:updready}
\end{figure}
\FloatBarrier
\newpage
\subsection*{Supplementary Note 5}
\label{section:loss-eval}

We have defined the task success criteria as MSE Loss $<0.1$ based on visual inspection. Below in Fig~\ref{fig:visualloss}, some network performances are shown.

\begin{figure}[!ht]
\centering
\begin{subfigure}{0.5\textwidth}
    \centering
    \includegraphics[width=1\textwidth]{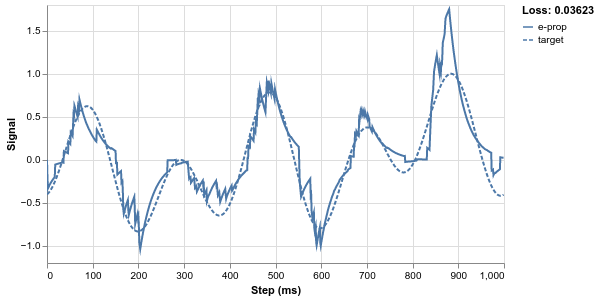}
\end{subfigure}%
\begin{subfigure}{0.5\textwidth}
    \centering
    \includegraphics[width=1\textwidth]{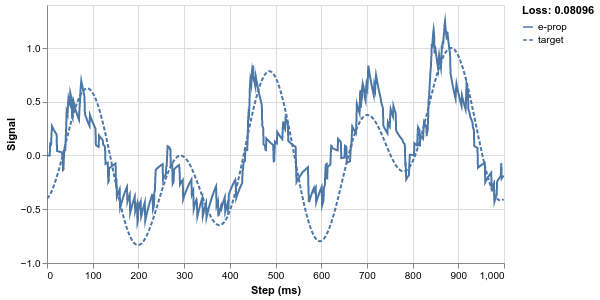}
\end{subfigure}
\begin{subfigure}{.5\textwidth}
    \centering
    \includegraphics[width=1\textwidth]{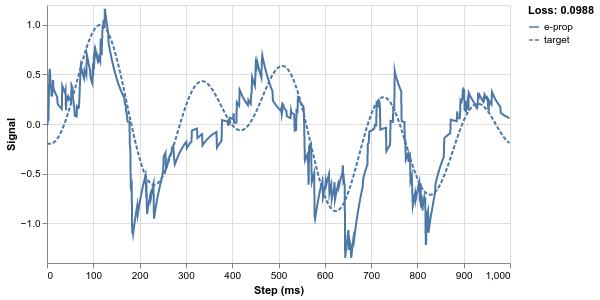}
\end{subfigure}%
\begin{subfigure}{.5\textwidth}
    \centering
    \includegraphics[width=1\textwidth]{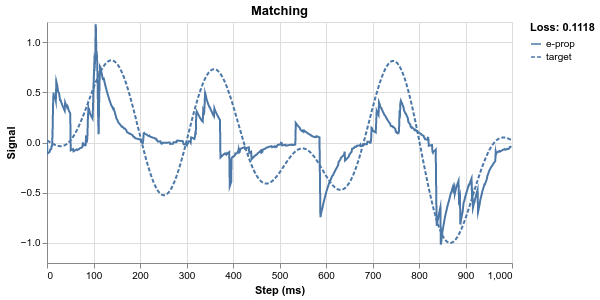}
\end{subfigure}
\begin{subfigure}{.5\textwidth}
    \centering
    \includegraphics[width=1\textwidth]{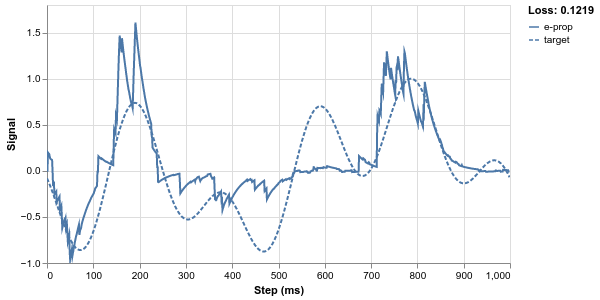}
\end{subfigure}%
\begin{subfigure}{.5\textwidth}
    \centering
    \includegraphics[width=1\textwidth]{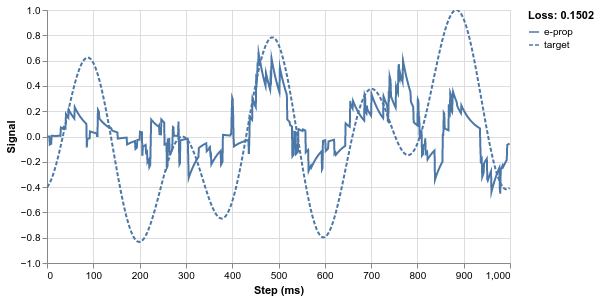}
\end{subfigure}
\caption[short]{Comparison of network performances with six different loss values.}
\label{fig:visualloss}
\end{figure}

\subsection*{Supplementary Note 6}
\label{section:networkstatsx}

\FloatBarrier

\begin{figure}[!ht]
  \centerfloat
  \includegraphics[width=0.85\textwidth]{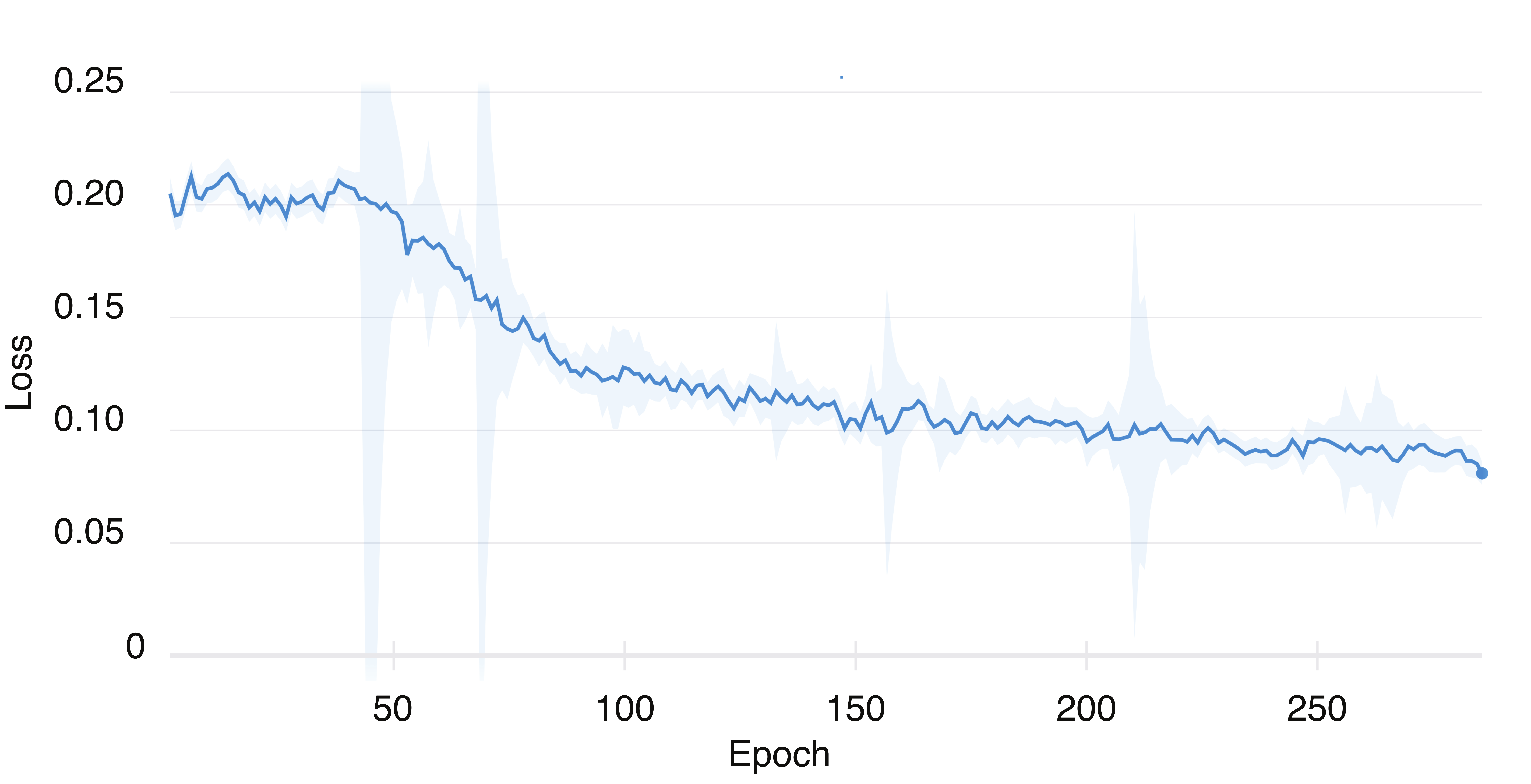}
  \caption{MSE loss of 50 networks trained with \ac{PCM} synapses using the mixed-precision method.}
  \label{fig:losstr}
\end{figure}
\FloatBarrier

\end{document}

%% file: acronyms.tex
\acrodef{ADC}[ADC]{Analog to Digital Converter}
\acrodef{ADEXP}[AdExp-I\&F]{Adaptive-Exponential Integrate and Fire}
\acrodef{AER}[AER]{Address-Event Representation}
\acrodef{AEX}[AEX]{AER EXtension board}
\acrodef{AE}[AE]{Address-Event}
\acrodef{AFM}[AFM]{Atomic Force Microscope}
\acrodef{AGC}[AGC]{Automatic Gain Control}
\acrodef{AI}[AI]{Artificial Intelligence}
\acrodef{ALD}[ALD]{Atomic Layer Deposition}
\acrodef{AMDA}[AMDA]{AER Motherboard with D/A converters}
\acrodef{ANN}[ANN]{Artificial Neural Network}
\acrodef{API}[API]{Application Programming Interface}
\acrodef{ARM}[ARM]{Advanced RISC Machine}
\acrodef{ASIC}[ASIC]{Application Specific Integrated Circuit}
\acrodef{AdExp}[AdExp-IF]{Adaptive Exponential Integrate-and-Fire}
\acrodef{BCI}[BCI]{Brain-Computer-Interface}
\acrodef{BCM}[BCM]{Bienenstock-Cooper-Munro}
\acrodef{BD}[BD]{Bundled Data}
\acrodef{BEOL}[BEOL]{Back-end of Line}
\acrodef{BG}[BG]{Bias Generator}
\acrodef{BMI}[BMI]{Brain-Machince Interface}
\acrodef{BTB}[BTB]{band-to-band tunnelling}
\acrodef{BP}[BP]{Back-propagation}
\acrodef{BPTT}[BPTT]{Back-propagation Through Time}
\acrodef{CAD}[CAD]{Computer Aided Design}
\acrodef{CAM}[CAM]{Content Addressable Memory}
\acrodef{CAVIAR}[CAVIAR]{Convolution AER Vision Architecture for Real-Time}
\acrodef{CA}[CA]{Cortical Automaton}
\acrodef{CCN}[CCN]{Cooperative and Competitive Network}
\acrodef{CDR}[CDR]{Clock-Data Recovery}
\acrodef{CFC}[CFC]{Current to Frequency Converter}
\acrodef{CHP}[CHP]{Communicating Hardware Processes}
\acrodef{CMIM}[CMIM]{Metal-insulator-metal Capacitor}
\acrodef{CML}[CML]{Current Mode Logic}
\acrodef{CMP}[CMP]{Chemical Mechanical Polishing}
\acrodef{CMOL}[CMOL]{Hybrid CMOS nanoelectronic circuits}
\acrodef{CMOS}[CMOS]{Complementary Metal-Oxide-Semiconductor}
\acrodef{CNN}[CCN]{Convolutional Neural Network}
\acrodef{COTS}[COTS]{Commercial Off-The-Shelf}
\acrodef{CPG}[CPG]{Central Pattern Generator}
\acrodef{CPLD}[CPLD]{Complex Programmable Logic Device}
\acrodef{CPU}[CPU]{Central Processing Unit}
\acrodef{CSM}[CSM]{Cortical State Machine}
\acrodef{CSP}[CSP]{Constraint Satisfaction Problem}
\acrodef{CV}[CV]{Coefficient of Variation}
\acrodef{DAC}[DAC]{Digital to Analog Converter}
\acrodef{DAS}[DAS]{Dynamic Auditory Sensor}
\acrodef{DAVIS}[DAVIS]{Dynamic and Active Pixel Vision Sensor}
\acrodef{DBN}[DBN]{Deep Belief Network}
\acrodef{DBS}[DBS]{Deep-Brain Stimulation}
\acrodef{DFA}[DFA]{Deterministic Finite Automaton}
\acrodef{DIBL}[DIBL]{drain-induced-barrier-lowering}
\acrodef{DI}[DI]{delay insensitive}
\acrodef{DMA}[DMA]{Direct Memory Access}
\acrodef{DNF}[DNF]{Dynamic Neural Field}
\acrodef{DNN}[DNN]{Deep Neural Network}
\acrodef{DoF}[DoF]{Degrees of Freedom}
\acrodef{DPE}[DPE]{Dynamic Parameter Estimation}
\acrodef{DPI}[DPI]{Differential Pair Integrator}
\acrodef{DRRZ}[DR-RZ]{Dual-Rail Return-to-Zero}
\acrodef{DRAM}[DRAM]{Dynamic Random Access Memory}
\acrodef{DR}[DR]{Dual Rail}
\acrodef{DSP}[DSP]{Digital Signal Processor}
\acrodef{DVS}[DVS]{Dynamic Vision Sensor}
\acrodef{DYNAP}[DYNAP]{Dynamic Neuromorphic Asynchronous Processor}
\acrodef{EBL}[EBL]{Electron Beam Lithography}
\acrodef{ECoG}[ECoG]{Electrocorticography}
\acrodef{EDVAC}[EDVAC]{Electronic Discrete Variable Automatic Computer}
\acrodef{EEG}[EEG]{electroencephalography}
\acrodef{EIN}[EIN]{Excitatory-Inhibitory Network}
\acrodef{EM}[EM]{Expectation Maximization}
\acrodef{EPSC}[EPSC]{Excitatory Post-Synaptic Current}
\acrodef{EPSP}[EPSP]{Excitatory Post-Synaptic Potential}
\acrodef{ET}[ET]{Eligibility Trace}
\acrodef{EZ}[EZ]{Epileptogenic Zone}
\acrodef{FDSOI}[FDSOI]{Fully-Depleted Silicon on Insulator}
\acrodef{FEOL}[FEOL]{Front-end of Line}
\acrodef{FET}[FET]{Field-Effect Transistor}
\acrodef{FFT}[FFT]{Fast Fourier Transform}
\acrodef{FI}[F-I]{Frequency-Current}
\acrodef{FPGA}[FPGA]{Field Programmable Gate Array}
\acrodef{FR}[FR]{Fast Ripple}
\acrodef{FSA}[FSA]{Finite State Automaton}
\acrodef{FSM}[FSM]{Finite State Machine}
\acrodef{GIDL}[GIDL]{gate-induced-drain-leakage}
\acrodef{GOPS}[GOPS]{Giga-Operations per Second}
\acrodef{GPU}[GPU]{Graphical Processing Unit}
\acrodef{GUI}[GUI]{Graphical User Interface}
\acrodef{HAL}[HAL]{Hardware Abstraction Layer}
\acrodef{HFO}[HFO]{High Frequency Oscillation}
\acrodef{HH}[H\&H]{Hodgkin \& Huxley}
\acrodef{HMM}[HMM]{Hidden Markov Model}
\acrodef{HRS}[HRS]{High-Resistive State}
\acrodef{HR}[HR]{Human Readable}
\acrodef{HSE}[HSE]{Handshaking Expansion}
\acrodef{HW}[HW]{Hardware}
\acrodef{IBCI}[IBCI]{Implantable BCI}
\acrodef{ICT}[ICT]{Information and Communication Technology}
\acrodef{IC}[IC]{Integrated Circuit}
\acrodef{ICL}[ICL]{Implantable Closed Loop}
\acrodef{IEEG}[iEEG]{intracranial electroencephalography}
\acrodef{IF2DWTA}[IF2DWTA]{Integrate \& Fire 2--Dimensional WTA}
\acrodef{IFSLWTA}[IFSLWTA]{Integrate \& Fire Stop Learning WTA}
\acrodef{IF}[I\&F]{Integrate-and-Fire}
\acrodef{IMU}[IMU]{Inertial Measurement Unit}
\acrodef{INCF}[INCF]{International Neuroinformatics Coordinating Facility}
\acrodef{INI}[INI]{Institute of Neuroinformatics}
\acrodef{INRC}[Intel NRC]{Intel Neuromorphic Research Community}
\acrodef{IO}[I/O]{Input/Output}
\acrodef{IoT}[IoT]{Internet of Things}
\acrodef{IPSC}[IPSC]{Inhibitory Post-Synaptic Current}
\acrodef{IPSP}[IPSP]{Inhibitory Post-Synaptic Potential}
\acrodef{IP}[IP]{Intellectual Property}
\acrodef{ISI}[ISI]{Inter-Spike Interval}
\acrodef{IoT}[IoT]{Internet of Things}
\acrodef{JFLAP}[JFLAP]{Java - Formal Languages and Automata Package}
\acrodef{LEDR}[LEDR]{Level-Encoded Dual-Rail}
\acrodef{LFP}[LFP]{Local Field Potential}
\acrodef{LLC}[LLC]{Low Leakage Cell}
\acrodef{LNA}[LNA]{Low-Noise Amplifier}
\acrodef{LPF}[LPF]{Low Pass Filter}
\acrodef{LRS}[LRS]{Low-Resistive State}
\acrodef{LSM}[LSM]{Liquid State Machine}
\acrodef{LSTM}[LSTM]{Long Short Term Memory}
\acrodef{LTD}[LTD]{Long Term Depression}
\acrodef{LTI}[LTI]{Linear Time-Invariant}
\acrodef{LTP}[LTP]{Long Term Potentiation}
\acrodef{LTU}[LTU]{Linear Threshold Unit}
\acrodef{LUT}[LUT]{Look-Up Table}
\acrodef{LVDS}[LVDS]{Low Voltage Differential Signaling}
\acrodef{MD}[MD]{Medical Device}
\acrodef{MCMC}[MCMC]{Markov-Chain Monte Carlo}
\acrodef{MEMS}[MEMS]{Micro Electro Mechanical System}
\acrodef{MFR}[MFR]{Mean Firing Rate}
\acrodef{MIM}[MIM]{Metal Insulator Metal}
\acrodef{ML}[ML]{Machine Leanring}
\acrodef{MLP}[MLP]{Multilayer Perceptron}
\acrodef{MOSCAP}[MOSCAP]{Metal Oxide Semiconductor Capacitor}
\acrodef{MOSFET}[MOSFET]{Metal Oxide Semiconductor Field-Effect Transistor}
\acrodef{MOS}[MOS]{Metal Oxide Semiconductor}
\acrodef{MRI}[MRI]{Magnetic Resonance Imaging}
\acrodef{NDFSM}[NDFSM]{Non-deterministic Finite State Machine} 
\acrodef{ND}[ND]{Noise-Driven}
\acrodef{NEF}[NEF]{Neural Engineering Framework}
\acrodef{NHML}[NHML]{Neuromorphic Hardware Mark-up Language}
\acrodef{NIL}[NIL]{Nano-Imprint Lithography}
\acrodef{NLP}[NLP]{Natural Language Processing}
\acrodef{NMDA}[NMDA]{N-Methyl-D-Aspartate}
\acrodef{NME}[NE]{Neuromorphic Engineering}
\acrodef{NN}[NN]{Neural Network}
\acrodef{NRZ}[NRZ]{Non-Return-to-Zero}
\acrodef{NSM}[NSM]{Neural State Machine}
\acrodef{OR}[OR]{Operating Room}
\acrodef{OTA}[OTA]{Operational Transconductance Amplifier}
\acrodef{PCB}[PCB]{Printed Circuit Board}
\acrodef{PCHB}[PCHB]{Pre-Charge Half-Buffer}
\acrodef{PCM}[PCM]{Phase Change Memory}
\acrodef{PD}[PD]{Parkinson Disease}
\acrodef{PE}[PE]{Phase Encoding}
\acrodef{PFA}[PFA]{Probabilistic Finite Automaton}
\acrodef{PFC}[PFC]{prefrontal cortex}
\acrodef{PFM}[PFM]{Pulse Frequency Modulation}
\acrodef{PM}[PM]{Personalized Medicine}
\acrodef{PR}[PR]{Production Rule}
\acrodef{PSC}[PSC]{Post-Synaptic Current}
\acrodef{PSP}[PSP]{Post-Synaptic Potential}
\acrodef{PSTH}[PSTH]{Peri-Stimulus Time Histogram}
\acrodef{PVD}[PVD]{Physical Vapor Deposition }
\acrodef{QDI}[QDI]{Quasi Delay Insensitive}
\acrodef{RAM}[RAM]{Random Access Memory}
\acrodef{RDF}[RDF]{random dopant fluctuation}
\acrodef{RELU}[ReLu]{Rectified Linear Unit}
\acrodef{RLS}[RLS]{Recursive Least-Squares}
\acrodef{RMSE}[RMSE]{Root Mean Squared-Error}
\acrodef{RMS}[RMS]{Root Mean Squared}
\acrodef{RNN}[RNN]{Recurrent Neural Network}
\acrodef{ROLLS}[ROLLS]{Reconfigurable On-Line Learning Spiking}
\acrodef{RRAM}[R-RAM]{Resistive Random Access Memory}
\acrodef{R}[R]{Ripples}
\acrodef{SAC}[SAC]{Selective Attention Chip}
\acrodef{SAT}[SAT]{Boolean Satisfiability Problem}
\acrodef{SCX}[SCX]{Silicon CorteX}
\acrodef{SD}[SD]{Signal-Driven}
\acrodef{SDSP}[SDSP]{Spike Driven Synaptic Plasticity}
\acrodef{SEM}[SEM]{Spike-based Expectation Maximization}
\acrodef{SLAM}[SLAM]{Simultaneous Localization and Mapping}
\acrodef{SNN}[SNN]{Spiking Neural Network}
\acrodef{SNR}[SNR]{Signal to Noise Ratio}
\acrodef{SOC}[SOC]{System-On-Chip}
\acrodef{SOI}[SOI]{Silicon on Insulator}
\acrodef{SoA}[SoA]{state-of-the-art}
\acrodef{SP}[SP]{Separation Property}
\acrodef{SRAM}[SRAM]{Static Random Access Memory}
\acrodef{STDP}[STDP]{Spike-Timing Dependent Plasticity}
\acrodef{STD}[STD]{Short-Term Depression}
\acrodef{STEM}[STEM]{Science, Technology, Engineering and Mathematics}
\acrodef{STP}[STP]{Short-Term Plasticity}
\acrodef{STT-MRAM}[STT-MRAM]{Spin-Transfer Torque Magnetic Random Access Memory}
\acrodef{STT}[STT]{Spin-Transfer Torque}
\acrodef{SW}[SW]{Software}
\acrodef{TCAM}[TCAM]{Ternary Content-Addressable Memory}
\acrodef{TFT}[TFT]{Thin Film Transistor}
\acrodef{TPU}[TPU]{Tensor Processing Unit}
\acrodef{TRL}[TRL]{Technology Readiness Level}
\acrodef{USB}[USB]{Universal Serial Bus}
\acrodef{VHDL}[VHDL]{VHSIC Hardware Description Language}
\acrodef{VLSI}[VLSI]{Very Large Scale Integration}
\acrodef{VOR}[VOR]{Vestibulo-Ocular Reflex}
\acrodef{WCST}[WCST]{Wisconsin Card Sorting Test}
\acrodef{WTA}[WTA]{Winner-Take-All}
\acrodef{XML}[XML]{eXtensible Mark-up Language}
\acrodef{CTXCTL}[CTXCTL]{CortexControl}
\acrodef{divmod3}[DIVMOD3]{divisibility of a number by three}
\acrodef{hWTA}[hWTA]{hard Winner-Take-All}
\acrodef{sWTA}[sWTA]{soft Winner-Take-All}
\acrodef{APMOM}[APMOM]{Alternate Polarity Metal On Metal}
\acrodef{GST}[GST]{$\text{Ge}_2\text{Sb}_2\text{Te}_5$}
\acrodef{LIF}[LIF]{Leaky Integrate-and-Fire}
\acrodef{RL}[RL]{Reinforcement Learning}
\acrodef{LB}[LB]{Learning Block}
\acrodef{OSTL}[OSTL]{Online Spatio-Temporal Learning}
\acrodef{RTRL}[RTRL]{Real Time Recurrent Learning}

%% file: main.bbl
\begin{thebibliography}{10}

\bibitem{Khrulkov_etal17}
Valentin Khrulkov, Alexander Novikov, and Ivan Oseledets.
\newblock {Expressive power of recurrent neural networks}.
\newblock {\em arXiv}, 2017.

\bibitem{Oord_etal16}
Aaron van~den Oord, Sander Dieleman, Heiga Zen, Karen Simonyan, Oriol Vinyals,
  Alex Graves, Nal Kalchbrenner, Andrew Senior, and Koray Kavukcuoglu.
\newblock {WaveNet: A Generative Model for Raw Audio}.
\newblock {\em arXiv:1609.03499 [cs]}, 09 2016.

\bibitem{Teed_Deng20}
Zachary Teed and Jia Deng.
\newblock {RAFT: Recurrent All-Pairs Field Transforms for Optical Flow}.
\newblock {\em arXiv}, 2020.

\bibitem{Brown_etal20}
Tom~B Brown, Benjamin Mann, Nick Ryder, Melanie Subbiah, Jared Kaplan, Prafulla
  Dhariwal, Arvind Neelakantan, Pranav Shyam, Girish Sastry, Amanda Askell,
  Sandhini Agarwal, Ariel Herbert-Voss, Gretchen Krueger, Tom Henighan, Rewon
  Child, Aditya Ramesh, Daniel~M Ziegler, Jeffrey Wu, Clemens Winter,
  Christopher Hesse, Mark Chen, Eric Sigler, Mateusz Litwin, Scott Gray,
  Benjamin Chess, Jack Clark, Christopher Berner, Sam McCandlish, Alec Radford,
  Ilya Sutskever, and Dario Amodei.
\newblock {Language Models are Few-Shot Learners}.
\newblock {\em arXiv}, 2020.

\bibitem{Berner_etal19}
Christopher Berner, Greg Brockman, Brooke Chan, Vicki Cheung, Christy Dennison,
  David Farhi, Quirin Fischer, Shariq Hashme, Chris Hesse, Rafal Józefowicz,
  Scott Gray, Catherine Olsson, Jakub Pachocki, Michael Petrov, Tim Salimans,
  Jeremy Schlatter, Jonas Schneider, Szymon Sidor, Ilya Sutskever, Jie Tang,
  Filip Wolski, and Susan Zhang.
\newblock {Dota 2 with Large Scale Deep Reinforcement Learning}.
\newblock page~66, 01 2019.

\bibitem{Ha_Schmidhuber18}
David Ha and Jürgen Schmidhuber.
\newblock {World Models}.
\newblock {\em arXiv:1803.10122 [cs, stat]}, 03 2018.

\bibitem{Vinyals_etal19}
Oriol Vinyals, Igor Babuschkin, Wojciech~M. Czarnecki, Michaël Mathieu, Andrew
  Dudzik, Junyoung Chung, David~H. Choi, Richard Powell, Timo Ewalds, Petko
  Georgiev, Junhyuk Oh, Dan Horgan, Manuel Kroiss, Ivo Danihelka, Aja Huang,
  Laurent Sifre, Trevor Cai, John~P. Agapiou, Max Jaderberg, Alexander~S.
  Vezhnevets, Rémi Leblond, Tobias Pohlen, Valentin Dalibard, David Budden,
  Yury Sulsky, James Molloy, Tom~L. Paine, Caglar Gulcehre, Ziyu Wang, Tobias
  Pfaff, Yuhuai Wu, Roman Ring, Dani Yogatama, Dario Wünsch, Katrina McKinney,
  Oliver Smith, Tom Schaul, Timothy Lillicrap, Koray Kavukcuoglu, Demis
  Hassabis, Chris Apps, and David Silver.
\newblock {Grandmaster level in StarCraft II using multi-agent reinforcement
  learning}.
\newblock {\em Nature}, 575(7782):350--354, 10 2019.

\bibitem{Douglas_Martin98}
R.J. Douglas and K.A.C. Martin.
\newblock Neocortex.
\newblock In G.M. Shepherd, editor, {\em The synaptic organization of the
  brain}, chapter~12, pages 459--509. Oxford University Press, Oxford, New
  York, 4th edition, 1998.

\bibitem{Douglas_etal95a}
R.J. Douglas, M.A. Mahowald, and C.~Mead.
\newblock Neuromorphic analogue {VLSI}.
\newblock {\em Annual Review of Neuroscience}, 18:255--281, 1995.

\bibitem{Zenke_Neftci21}
Friedemann Zenke and Emre~O Neftci.
\newblock {Brain-Inspired Learning on Neuromorphic Substrates}.
\newblock {\em Proceedings of the IEEE}, pages 1--16, 2021.

\bibitem{Ambrogio_etal18}
Stefano Ambrogio, Pritish Narayanan, Hsinyu Tsai, Robert~M. Shelby, Irem
  Boybat, Carmelo di~Nolfo, Severin Sidler, Massimo Giordano, Martina Bodini,
  Nathan C.~P. Farinha, Benjamin Killeen, Christina Cheng, Yassine Jaoudi, and
  Geoffrey~W. Burr.
\newblock Equivalent-accuracy accelerated neural-network training using
  analogue memory.
\newblock {\em Nature}, 558(7708):60--67, June 2018.

\bibitem{Li_etal18a}
Can Li, Daniel Belkin, Yunning Li, Peng Yan, Miao Hu, Ning Ge, Hao Jiang, Eric
  Montgomery, Peng Lin, Zhongrui Wang, Wenhao Song, John~Paul Strachan, Mark
  Barnell, Qing Wu, R.~Stanley Williams, J.~Joshua Yang, and Qiangfei Xia.
\newblock Efficient and self-adaptive in-situ learning in multilayer memristor
  neural network.
\newblock {\em Nature {C}ommunications}, 9(2385):1--8, June 2018.

\bibitem{Prezioso_etal15}
Mirko Prezioso, Farnood Merrikh-Bayat, BD~Hoskins, GC~Adam, Konstantin~K
  Likharev, and Dmitri~B Strukov.
\newblock Training and operation of an integrated neuromorphic network based on
  metal-oxide memristors.
\newblock {\em Nature}, 521(7550):61--64, 2015.

\bibitem{dalgaty_etal21}
Thomas Dalgaty, Niccolo Castellani, Cl{\'e}ment Turck, Kamel-Eddine Harabi,
  Damien Querlioz, and Elisa Vianello.
\newblock In situ learning using intrinsic memristor variability via markov
  chain monte carlo sampling.
\newblock {\em Nature Electronics}, 4(2):151--161, 2021.

\bibitem{cai_etal20}
Fuxi Cai, Suhas Kumar, Thomas Van~Vaerenbergh, Xia Sheng, Rui Liu, Can Li, Zhan
  Liu, Martin Foltin, Shimeng Yu, Qiangfei Xia, et~al.
\newblock Power-efficient combinatorial optimization using intrinsic noise in
  memristor hopfield neural networks.
\newblock {\em Nature Electronics}, 3(7):409--418, 2020.

\bibitem{Sebastian_etal19}
Abu Sebastian, Manuel~Le Gallo, and Evangelos Eleftheriou.
\newblock Computational phase-change memory: beyond {von Neumann} computing.
\newblock {\em Journal of Physics D: Applied Physics}, 52(44):443002, August
  2019.

\bibitem{Backus78}
J.~Backus.
\newblock Can programming be liberated from the {von Neumann} style?: a
  functional style and its algebra of programs.
\newblock {\em Communications of the ACM}, 21(8):613--641, 1978.

\bibitem{Indiveri_Liu15}
G.~Indiveri and S.-C. Liu.
\newblock Memory and information processing in neuromorphic systems.
\newblock {\em Proceedings of the {IEEE}}, 103(8):1379--1397, 2015.

\bibitem{Payvand_etal19}
Melika Payvand, Manu~V Nair, Lorenz~K M{\"u}ller, and Giacomo Indiveri.
\newblock A neuromorphic systems approach to in-memory computing with non-ideal
  memristive devices: From mitigation to exploitation.
\newblock {\em Faraday Discussions}, 213:487--510, 2019.

\bibitem{Chicca_Indiveri20}
Elisabetta Chicca and Giacomo Indiveri.
\newblock A recipe for creating ideal hybrid memristive-{CMOS} neuromorphic
  processing systems.
\newblock {\em Applied Physics Letters}, 116(12):120501, 2020.

\bibitem{Peng_etal19}
Xiaochen Peng, Shanshi Huang, Yandong Luo, Xiaoyu Sun, and Shimeng Yu.
\newblock {DNN+NeuroSim: An End-to-End Benchmarking Framework for Compute
  -in-Memory Accelerators with Versatile Device Technologies}.
\newblock {\em IEEE International Electron Devices Meeting (IEDM)}, pages
  32.5.1--32.5.4, 2019.

\bibitem{Peng_etal20}
Xiaochen Peng, Shanshi Huang, Hongwu Jiang, Anni Lu, and Shimeng Yu.
\newblock {DNN+NeuroSim V2.0: An End-to-End Benchmarking Framework for
  Compute-in-Memory Accelerators for On-chip Training}.
\newblock {\em IEEE Transactions on Computer-Aided Design of Integrated
  Circuits and Systems}, PP(99):1--1, 2020.

\bibitem{Burr_etal16a}
Geoffrey~W. Burr, Matthew~J. Brightsky, Abu Sebastian, Huai-Yu Cheng, Jau-Yi
  Wu, Sangbum Kim, Norma~E. Sosa, Nikolaos Papandreou, Hsiang-Lan Lung,
  Haralampos Pozidis, Evangelos Eleftheriou, and Chung~H. Lam.
\newblock {Recent Progress in Phase-Change Memory Technology}.
\newblock {\em IEEE Journal on Emerging and Selected Topics in Circuits and
  Systems}, 6(2):146--162, 06 2016.

\bibitem{Burr_etal17}
Geoffrey~W Burr, Robert~M Shelby, Abu Sebastian, Sangbum Kim, Seyoung Kim,
  Severin Sidler, Kumar Virwani, Masatoshi Ishii, Pritish Narayanan, Alessandro
  Fumarola, Lucas~L Sanches, Irem Boybat, Manuel Le~Gallo, Kibong Moon, Jiyoo
  Woo, Hyunsang Hwang, and Yusuf Leblebici.
\newblock Neuromorphic computing using non-volatile memory.
\newblock {\em Advances in Physics: X}, 2(1):89--124, 2017.

\bibitem{Tuma_etal16}
Tomas Tuma, Angeliki Pantazi, Manuel Le~Gallo, Abu Sebastian, and Evangelos
  Eleftheriou.
\newblock Stochastic phase-change neurons.
\newblock {\em Nature Nanotechnology}, 11(8):693--699, 2016.

\bibitem{Karunaratne_etal20a}
Geethan Karunaratne, Manuel~Le Gallo, Giovanni Cherubini, Luca Benini, Abbas
  Rahimi, and Abu Sebastian.
\newblock {In-memory hyperdimensional computing}.
\newblock {\em Nature Electronics}, 3(6):327--337, 06 2020.

\bibitem{Demirag_etal21a}
Yigit Demirag, Filippo Moro, Thomas Dalgaty, Gabriele Navarro, Charlotte
  Frenkel, Giacomo Indiveri, Elisa Vianello, and Melika Payvand.
\newblock {PCM-trace: Scalable Synaptic Eligibility Traces with Resistivity
  Drift of Phase-Change Materials}.
\newblock {\em 2021 IEEE International Symposium on Circuits and Systems
  (ISCAS)}, pages 1--5, 2021.

\bibitem{Gallo_etal18}
Manuel~Le Gallo, Abu Sebastian, Giovanni Cherubini, Heiner Giefers, and
  Evangelos Eleftheriou.
\newblock {Compressed Sensing With Approximate Message Passing Using In-Memory
  Computing}.
\newblock {\em IEEE Transactions on Electron Devices}, 65(10):4304--4312, 2018.

\bibitem{Nandakumar_etal18}
SR~Nandakumar, Manuel Le~Gallo, Irem Boybat, Bipin Rajendran, Abu Sebastian,
  and Evangelos Eleftheriou.
\newblock A phase-change memory model for neuromorphic computing.
\newblock {\em Journal of Applied Physics}, 124(15):152135, 2018.

\bibitem{Bellec_etal20}
Guillaume Bellec, Franz Scherr, Anand Subramoney, Elias Hajek, Darjan Salaj,
  Robert Legenstein, and Wolfgang Maass.
\newblock A solution to the learning dilemma for recurrent networks of spiking
  neurons.
\newblock {\em Nature Communications}, 11(3625):1--15, 2020.

\bibitem{Demirag18}
Yigit Demirag.
\newblock Multiphysics modeling of {$\text{Ge}_2\text{Sb}_2\text{Te}_5$} based
  synaptic devices for brain inspired computing.
\newblock Master's thesis, Ihsan Dogramaci Bilkent University, Ankara, Turkey,
  July 2018.

\bibitem{Gallo_Sebastian20}
Manuel~Le Gallo and Abu Sebastian.
\newblock {An overview of phase-change memory device physics}.
\newblock {\em Journal of Physics D: Applied Physics}, 53(21):213002, 2020.

\bibitem{Gallo_etal15}
Manuel~Le Gallo, Matthias Kaes, Abu Sebastian, and Daniel Krebs.
\newblock {Subthreshold electrical transport in amorphous phase-change
  materials}.
\newblock {\em New Journal of Physics}, 17(9):093035, 09 2015.

\bibitem{Gallo_etal16}
Manuel~Le Gallo, Aravinthan Athmanathan, Daniel Krebs, and Abu Sebastian.
\newblock {Evidence for thermally assisted threshold switching behavior in
  nanoscale phase-change memory cells}.
\newblock {\em Journal of Applied Physics}, 119(2):025704, 01 2016.

\bibitem{Ielmini_etal07}
D.~Ielmini, S.~Lavizzari, D.~Sharma, and A.~L. Lacaita.
\newblock {Physical Interpretation, Modeling and Impact on Phase Change Memory
  (PCM) Reliability of Resistance Drift Due to Chalcogenide Structural
  Relaxation}.
\newblock {\em 2007 IEEE International Electron Devices Meeting}, pages
  939--942, 2007.

\bibitem{Karpov_etal07}
IV~Karpov, M~Mitra, D~Kau, G~Spadini, YA~Kryukov, and VG~Karpov.
\newblock Fundamental drift of parameters in chalcogenide phase change memory.
\newblock {\em Journal of Applied Physics}, 102(12):124503, 2007.

\bibitem{Redaelli_etal08}
A.~Redaelli, A.~Pirovano, A.~Benvenuti, and A.~L. Lacaita.
\newblock {Threshold switching and phase transition numerical models for phase
  change memory simulations}.
\newblock {\em Journal of Applied Physics}, 103(11):111101, 2008.

\bibitem{Salinga_etal13}
Martin Salinga, Egidio Carria, Andreas Kaldenbach, Manuel Bornhöfft, Julia
  Benke, Joachim Mayer, and Matthias Wuttig.
\newblock {Measurement of crystal growth velocity in a melt-quenched
  phase-change material}.
\newblock {\em Nature Communications}, 4(1):2371, 2013.

\bibitem{Nardone_etal09}
M~Nardone, V~I Kozub, I~V Karpov, and V~G Karpov.
\newblock {Possible mechanisms for 1/f noise in chalcogenide glasses: A
  theoretical description}.
\newblock {\em Physical Review B}, 79(16):165206, 2009.

\bibitem{Paszke_etal19a}
Adam Paszke, Sam Gross, Francisco Massa, Adam Lerer, James Bradbury, Gregory
  Chanan, Trevor Killeen, Zeming Lin, Natalia Gimelshein, Luca Antiga, Alban
  Desmaison, Andreas Kopf, Edward Yang, Zachary DeVito, Martin Raison, Alykhan
  Tejani, Sasank Chilamkurthy, Benoit Steiner, Lu~Fang, Junjie Bai, and Soumith
  Chintala.
\newblock Pytorch: An imperative style, high-performance deep learning library.
\newblock In H.~Wallach, H.~Larochelle, A.~Beygelzimer, F.~d\textquotesingle
  Alch\'{e}-Buc, E.~Fox, and R.~Garnett, editors, {\em Advances in Neural
  Information Processing Systems 32}, pages 8024--8035. Curran Associates,
  Inc., 2019.

\bibitem{Richards_etal19}
Blake~A. Richards, Timothy~P. Lillicrap, Philippe Beaudoin, Yoshua Bengio,
  Rafal Bogacz, Amelia Christensen, Claudia Clopath, Rui~Ponte Costa, Archy~de
  Berker, Surya Ganguli, Colleen~J. Gillon, Danijar Hafner, Adam Kepecs,
  Nikolaus Kriegeskorte, Peter Latham, Grace~W. Lindsay, Kenneth~D. Miller,
  Richard Naud, Christopher~C. Pack, Panayiota Poirazi, Pieter Roelfsema, João
  Sacramento, Andrew Saxe, Benjamin Scellier, Anna~C. Schapiro, Walter Senn,
  Greg Wayne, Daniel Yamins, Friedemann Zenke, Joel Zylberberg, Denis Therien,
  and Konrad~P. Kording.
\newblock {A deep learning framework for neuroscience}.
\newblock {\em Nature Neuroscience}, 22(11):1761--1770, 11 2019.

\bibitem{Fremaux_Gerstner16}
Nicolas Fr{\'e}maux and Wulfram Gerstner.
\newblock Neuromodulated spike-timing-dependent plasticity, and theory of
  three-factor learning rules.
\newblock {\em Front. Neur. Circ.}, 9:85, 2016.

\bibitem{Sacramento_etal18}
Jo{\~a}o Sacramento, Rui~Ponte Costa, Yoshua Bengio, and Walter Senn.
\newblock Dendritic cortical microcircuits approximate the backpropagation
  algorithm.
\newblock In {\em Advances in Neural Information Processing Systems}, pages
  8721--8732, 2018.

\bibitem{Pozzi_etal18}
Isabella Pozzi, Sander Bohté, and Pieter Roelfsema.
\newblock {A Biologically Plausible Learning Rule for Deep Learning in the
  Brain}.
\newblock {\em arXiv}, 2018.

\bibitem{Gerstner_Kistler02}
Wulfram Gerstner and Werner~M Kistler.
\newblock {\em Spiking neuron models: Single neurons, populations, plasticity}.
\newblock Cambridge university press, Cambridge, United Kingdom, 2002.

\bibitem{Dayan_Abbott01}
P.~Dayan and L.F. Abbott.
\newblock {\em Theoretical Neuroscience: Computational and Mathematical
  Modeling of Neural Systems}.
\newblock MIT Press, Cambridge, MA, USA, 2001.

\bibitem{Sussillo_Abbott09}
D.~Sussillo and L.F. Abbott.
\newblock Generating coherent patterns of activity from chaotic neural
  networks.
\newblock {\em Neuron}, 63(4):544--557, 2009.

\bibitem{Diehl_etal15}
Peter~U Diehl, Daniel Neil, Jonathan Binas, Matthew Cook, Shih-Chii Liu, and
  Michael Pfeiffer.
\newblock Fast-classifying, high-accuracy spiking deep networks through weight
  and threshold balancing.
\newblock In {\em International Joint Conference on Neural Networks ({IJCNN})},
  pages 1--8. IEEE, 2015.

\bibitem{Nicola_Clopath17a}
Wilten Nicola and Claudia Clopath.
\newblock {Supervised Learning in Spiking Neural Networks with FORCE Training}.
\newblock {\em Nature Communications}, 8(1):2208, 12 2017.

\bibitem{Neftci_etal19}
Emre~O Neftci, Hesham Mostafa, and Friedemann Zenke.
\newblock Surrogate gradient learning in spiking neural networks: Bringing the
  power of gradient-based optimization to spiking neural networks.
\newblock {\em {IEEE} Signal Processing Magazine}, 36(6):51--63, 2019.

\bibitem{Lee_etal16}
Jun~Haeng Lee, Tobi Delbruck, and Michael Pfeiffer.
\newblock Training deep spiking neural networks using backpropagation.
\newblock {\em Frontiers in Neuroscience}, 10:508, 2016.

\bibitem{Hochreiter_Schmidhuber97}
Sepp Hochreiter and J{\"u}rgen Schmidhuber.
\newblock Long short-term memory.
\newblock {\em Neural computation}, 9(8):1735--1780, 1997.

\bibitem{Bohnstingl_etal20}
Thomas Bohnstingl, Stanisław Woźniak, Wolfgang Maass, Angeliki Pantazi, and
  Evangelos Eleftheriou.
\newblock {Online Spatio-Temporal Learning in Deep Neural Networks}.
\newblock {\em arXiv}, 2020.

\bibitem{Qiao_etal15}
Ning Qiao, Hesham Mostafa, Federico Corradi, Marc Osswald, Fabio Stefanini,
  Dora Sumislawska, and Giacomo Indiveri.
\newblock A reconfigurable on-line learning spiking neuromorphic processor
  comprising 256 neurons and 128k synapses.
\newblock {\em Frontiers in neuroscience}, 9:141, 2015.

\bibitem{Payvand_etal18}
Melika Payvand, Lorenz~K Muller, and Giacomo Indiveri.
\newblock Event-based circuits for controlling stochastic learning with
  memristive devices in neuromorphic architectures.
\newblock In {\em Circuits and Systems (ISCAS), 2018 IEEE International
  Symposium on}, pages 1--5. IEEE, 2018.

\bibitem{Payvand_etal20}
Melika Payvand, Mohammed~E. Fouda, Fadi Kurdahi, Ahmed~M. Eltawil, and Emre~O.
  Neftci.
\newblock {On-Chip Error-Triggered Learning of Multi-Layer Memristive Spiking
  Neural Networks}.
\newblock {\em IEEE Journal on Emerging and Selected Topics in Circuits and
  Systems}, 10(4):522--535, 2020.

\bibitem{Boybat_etal18}
Irem Boybat, Manuel~Le Gallo, Timoleon Moraitis, Thomas Parnell, Tomas Tuma,
  Bipin Rajendran, Yusuf Leblebici, Abu Sebastian, and Evangelos Eleftheriou.
\newblock Neuromorphic computing with multi-memristive synapses.
\newblock {\em Nature {C}ommunications}, 9:2514, 2018.

\bibitem{Nandakumar_etal20a}
S.~R. Nandakumar, Manuel~Le Gallo, Christophe Piveteau, Vinay Joshi, Giovanni
  Mariani, Irem Boybat, Geethan Karunaratne, Riduan Khaddam-Aljameh, Urs Egger,
  Anastasios Petropoulos, Theodore Antonakopoulos, Bipin Rajendran, Abu
  Sebastian, and Evangelos Eleftheriou.
\newblock {Mixed-Precision Deep Learning Based on Computational Memory}.
\newblock {\em Frontiers in Neuroscience}, 14:406, 2020.

\bibitem{Nair_etal17}
Manu~V Nair, Lorenz~K. Mueller, and Giacomo Indiveri.
\newblock A differential memristive synapse circuit for on-line learning in
  neuromorphic computing systems.
\newblock {\em Nano Futures}, 1(3):1--12, 2017.

\bibitem{Burr_etal14}
G.W. Burr, R.M. Shelby, C.di Nolfo, J.W. Jang, R.S. Shenoy, P.~Narayanan,
  K.~Virwani, E.U. Giacometti, B.~Kurdi, and H.~Hwang.
\newblock {Experimental Demonstration and Tolerancing of a Large-Scale Neural
  Network (165,000 Synapses), using Phase-Change Memory as the Synaptic Weight
  Element}.
\newblock {\em 2014 IEEE International Electron Devices Meeting}, pages 1--4,
  2014.

\bibitem{Balles_etal20a}
Lukas Balles, Fabian Pedregosa, and Nicolas~Le Roux.
\newblock {The Geometry of Sign Gradient Descent}.
\newblock {\em arXiv}, 2020.

\bibitem{Nair_Dudek15}
Manu~V Nair and Piotr Dudek.
\newblock Gradient-descent-based learning in memristive crossbar arrays.
\newblock In {\em International Joint Conference on Neural Networks ({IJCNN})},
  pages 1--7. IEEE, 2015.

\bibitem{Muller_etal17}
L.~M{\"u}ller, M.~Nair, and G.~Indiveri.
\newblock Randomized unregulated step descent for limited precision synaptic
  elements.
\newblock In {\em International Symposium on Circuits and Systems, ({ISCAS})}.
  IEEE, 2017.

\bibitem{Payvand_etal20a}
Melika Payvand, Mohammed~E. Fouda, Fadi Kurdahi, Ahmed~M. Eltawil, and Emre~O.
  Neftci.
\newblock {On-Chip Error-Triggered Learning of Multi-Layer Memristive Spiking
  Neural Networks}.
\newblock {\em IEEE Journal on Emerging and Selected Topics in Circuits and
  Systems}, 10(4):522--535, 2020.

\bibitem{Kingma_Ba14}
Diederik~P Kingma and Jimmy Ba.
\newblock Adam: A method for stochastic optimization.
\newblock {\em arXiv preprint arXiv:1412.6980}, 2014.

\bibitem{Athmanathan_etal16a}
Aravinthan Athmanathan, Milos Stanisavljevic, Nikolaos Papandreou, Haralampos
  Pozidis, and Evangelos Eleftheriou.
\newblock {Multilevel-Cell Phase-Change Memory: A Viable Technology}.
\newblock {\em IEEE Journal on Emerging and Selected Topics in Circuits and
  Systems}, 6(1):87--100, 2016.

\bibitem{Hubara_etal16}
Itay Hubara, Matthieu Courbariaux, Daniel Soudry, Ran El-Yaniv, and Yoshua
  Bengio.
\newblock {Quantized Neural Networks: Training Neural Networks with Low
  Precision Weights and Activations}.
\newblock {\em arXiv}, 2016.

\bibitem{davies_etal21}
Mike Davies, Andreas Wild, Garrick Orchard, Yulia Sandamirskaya, Gabriel
  A~Fonseca Guerra, Prasad Joshi, Philipp Plank, and Sumedh~R Risbud.
\newblock Advancing neuromorphic computing with loihi: A survey of results and
  outlook.
\newblock {\em Proceedings of the IEEE}, 109(5):911--934, 2021.

\bibitem{frenkel_etal21}
Charlotte Frenkel, David Bol, and Giacomo Indiveri.
\newblock Bottom-up and top-down neural processing systems design: Neuromorphic
  intelligence as the convergence of natural and artificial intelligence.
\newblock {\em arXiv preprint arXiv:2106.01288}, 2021.

\bibitem{Fusi_Abbott07}
S.~Fusi and L.F. Abbott.
\newblock Limits on the memory storage capacity of bounded synapses.
\newblock {\em Nature Neuroscience}, 10:485--493, 2007.

\bibitem{laborieux_etal21}
Axel Laborieux, Maxence Ernoult, Tifenn Hirtzlin, and Damien Querlioz.
\newblock Synaptic metaplasticity in binarized neural networks.
\newblock {\em Nature communications}, 12(1):1--12, 2021.

\bibitem{Muller_Indiveri15a}
Lorenz~K Muller and Giacomo Indiveri.
\newblock Rounding methods for neural networks with low resolution synaptic
  weights.
\newblock {\em arXiv preprint arXiv:1504.05767}, pages 1--11, 2015.

\bibitem{Frenkel_etal19}
Charlotte Frenkel, Jean-Didier Legat, and David Bol.
\newblock Morphic: A 65-nm 738k-synapse/mm$^2$ quad-core binary-weight digital
  neuromorphic processor with stochastic spike-driven online learning.
\newblock {\em IEEE Transactions on Biomedical Circuits and Systems},
  13(5):999--1010, 2019.

\bibitem{Frenkel_etal20}
Charlotte Frenkel, Jean-Didier Legat, and David Bol.
\newblock A 28-nm convolutional neuromorphic processor enabling online learning
  with spike-based retinas.
\newblock In {\em 2020 IEEE International Symposium on Circuits and Systems
  (ISCAS)}, pages 1--5. IEEE, 2020.

\bibitem{Williams_Zipser89}
Ronald~J. Williams and David Zipser.
\newblock {A Learning Algorithm for Continually Running Fully Recurrent Neural
  Networks}.
\newblock {\em Neural Computation}, 1(2):270--280, 1989.

\bibitem{Bellec_etal18}
Guillaume Bellec, Darjan Salaj, Anand Subramoney, Robert Legenstein, and
  Wolfgang Maass.
\newblock Long short-term memory and learning-to-learn in networks of spiking
  neurons.
\newblock In {\em Advances in Neural Information Processing Systems}, pages
  787--797, 2018.

\bibitem{Snoek_etal12a}
Jasper Snoek, Hugo Larochelle, and Ryan~P. Adams.
\newblock Practical bayesian optimization of machine learning algorithms.
\newblock In {\em Proceedings of the 25th International Conference on Neural
  Information Processing Systems - Volume 2}, NIPS'12, page 2951–2959, Red
  Hook, NY, USA, 2012. Curran Associates Inc.

\end{thebibliography}
